\documentclass{article}

\usepackage{amsmath,amsfonts,bm}

\def\eqref#1{equation~\ref{#1}}

\def\1{\bm{1}}

\def\rvp{{\mathbf{p}}}

\def\rvy{{\mathbf{y}}}

\def\rmA{{\mathbf{A}}}

\def\rmX{{\mathbf{X}}}

\DeclareMathAlphabet{\mathsfit}{\encodingdefault}{\sfdefault}{m}{sl}
\SetMathAlphabet{\mathsfit}{bold}{\encodingdefault}{\sfdefault}{bx}{n}

\def\sD{{\mathbb{D}}}

\newcommand{\E}{\mathbb{E}}

\newcommand{\KL}{D_{\mathrm{KL}}}
\newcommand{\Var}{\mathrm{Var}}

\usepackage[accepted]{icml2021}
\usepackage{times}  
\usepackage{helvet} 
\usepackage{courier}  
\usepackage[hyphens]{url}  
\usepackage{algorithm}
\usepackage{algorithmic}
\renewcommand{\algorithmicensure}{ \textbf{Output:}} 
\renewcommand{\algorithmicrequire}{ \textbf{Input:}}
\usepackage{hyperref}
\hypersetup{
     colorlinks=true,
     linkcolor=blue,
     filecolor=blue,
     citecolor = blue,      
     urlcolor=cyan,
 }
\usepackage{microtype}
\usepackage{graphicx}
\usepackage{subfigure}
\usepackage{booktabs} 
\usepackage{amsmath}
\usepackage{amssymb}
\usepackage{booktabs}
\usepackage{multirow}
\usepackage{subfigure}
\usepackage{makecell}
\usepackage{xcolor}

\icmltitlerunning{KD3A: Unsupervised Multi-Source Decentralized Domain Adaptation via Knowledge Distillation}

\begin{document}

\twocolumn[
\icmltitle{KD3A: Unsupervised Multi-Source Decentralized Domain Adaptation via Knowledge Distillation}




\begin{icmlauthorlist}
\icmlauthor{Hao-zhe Feng}{ZJU-CAD}
\icmlauthor{Zhaoyang You}{ZJU-CS}
\icmlauthor{Minghao Chen}{ZJU-CAD}
\icmlauthor{Tianye Zhang}{ZJU-CAD}
\icmlauthor{Minfeng Zhu}{ZJU-CAD}\\
\icmlauthor{Fei Wu}{ZJU-CS}
\icmlauthor{Chao Wu}{ZJU-SPA}
\icmlauthor{Wei Chen}{ZJU-CAD}
\end{icmlauthorlist}

\icmlcorrespondingauthor{Wei Chen}{chenvis@zju.edu.cn}
\icmlaffiliation{ZJU-CAD}{State Key Lab of CAD\&CG, Zhejiang University, Hangzhou, China}
\icmlaffiliation{ZJU-CS}{College of Computer Science and Technology, Zhejiang University, Hangzhou, China}
\icmlaffiliation{ZJU-SPA}{School of Public Affairs, Zhejiang University, Hangzhou, China}
\icmlkeywords{Federated Learning, Domain Adaptation, Self-supervised Learning}

\vskip 0.3in
]



\printAffiliationsAndNotice{} 

\begin{abstract}
Conventional unsupervised multi-source domain adaptation (UMDA) methods assume all source domains can be accessed directly. However, this assumption neglects the privacy-preserving policy, where all the data and computations must be kept decentralized. There exist three challenges in this scenario: (1) Minimizing the domain distance requires the pairwise calculation of the data from source and target domains, while the data on the source domain is not available. (2) The communication cost and privacy security limit the application of existing UMDA methods, such as the domain adversarial training. (3) Since users cannot govern the data quality, the irrelevant or malicious source domains are more likely to appear, which causes negative transfer. To address the above problems, we propose a privacy-preserving UMDA paradigm named \textbf{K}nowledge \textbf{D}istillation based \textbf{D}ecentralized \textbf{D}omain \textbf{A}daptation (KD3A), which performs domain adaptation through the knowledge distillation on models from different source domains. The extensive experiments show that KD3A significantly outperforms state-of-the-art UMDA approaches. Moreover, the KD3A is robust to the negative transfer and brings a 100$\times$ reduction of communication cost compared with other decentralized UMDA methods. 
\end{abstract}

\section{Introduction}\label{sec:introduction}
Most deep learning models are trained with large-scale datasets via supervised learning. Since it is often costly to get sufficient data, we usually use other similar datasets to train the model. However, due to the domain shift, naively combining different datasets often results in unsatisfying performance. \textbf{U}nsupervised \textbf{M}ulti-source \textbf{D}omain \textbf{A}daptation (UMDA) \citep{DBLP:conf/aaai/ZhangGS15} addresses such problems by establishing transferable features from multiple source domains to an unlabeled target domain.

Recent advanced UMDA methods \cite{DBLP:conf/cvpr/ChangYSKH19,DBLP:conf/aaai/ZhaoWZGLS0HCK20} perform the knowledge transfer within two steps: (1) Combining data from source and target domains to construct \textbf{S}ource-\textbf{T}arget pairs. (2) Establishing transferable features by minimizing the $\mathcal{H}$-divergence. This prevailing paradigm works well when all source domains are available. However, in terms of the privacy-preserving policy, we cannot access the sensitive data such as the patient data from different hospitals and the client profiles from different companies. In these cases, all the data and computations on source domains must be kept decentralized.

Most conventional UMDA methods are not applicable under this privacy-preserving policy due to three problems: (1) Minimizing the $\mathcal{H}$-divergence in UMDA requires the pairwise calculation of the data from source and target domains, while the data on source domain is not available. (2) The communication cost and privacy security limit the application of advanced UMDA methods. For example, the domain adversarial training is able to optimize the $\mathcal{H}$-divergence without accessing data \citep{DBLP:conf/iclr/PengHZS20}. However, it requires each source domain to synchronize model with target domain after every single batch, which results in huge communication cost and causes the privacy leakage \citep{DBLP:conf/nips/ZhuLH19}. (3) The negative transfer problem \citep{5288526}. Since it is difficult to govern the data quality, there can exist some irrelevant source domains that are very different from the target domain or even some malicious source domains which perform the poisoning attack \cite{DBLP:conf/aistats/BagdasaryanVHES20}. With these bad domains, the negative transfer occurs.

In this study, we propose a solution to the above problems, \textbf{K}nowledge \textbf{D}istillation based \textbf{D}ecentralized \textbf{D}omain \textbf{A}daptation (KD3A), which aims to perform decentralized domain adaptation through the knowledge distillation on  models from different source domains. Our KD3A approach consists of three components used in tandem. First, we propose a multi-source knowledge distillation method named \textit{Knowledge Vote} to obtain high-quality domain consensus. Based on the consensus quality of different source domains, we devise a dynamic weighting strategy named \textit{Consensus Focus} to identify the malicious and irrelevant source domains. Finally, we derive a decentralized optimization strategy of $\mathcal{H}$-divergence named \textit{BatchNorm MMD}. Moreover, we analyze the decentralized generalization bound for KD3A from a theoretical perspective. The extensive experiments show our KD3A has the following advantages:
\begin{itemize}
    \item The KD3A brings a 100$\times$ reduction of communication cost compared with other decentralized UMDA methods and is robust to the privacy leakage attack.
    \item The KD3A assigns low weights to those malicious or irrelevant domains. Therefore, it is robust to negative transfer.
    \item The KD3A significantly outperforms the \textit{state-of-the-art} UMDA approaches with $51.1\%$ accuracy on the large-scale DomainNet dataset. 
\end{itemize}
In addition, our KD3A is easy to implement and we create an \href{https://github.com/FengHZ/KD3A}{open-source framework} to conduct KD3A on different benchmarks.
\section{Related work}
\subsection{Unsupervised Multi-Source Domain Adaptation}
Unsupervised Multi-source Domain Adaptation (UMDA) establish the transferable features by reducing the $\mathcal{H}$-divergence between source domain $\sD_{S}$ and target domain $\sD_{T}$. There are two prevailing paradigms that provide the optimization strategy of $\mathcal{H}$-divergence, i.e. maximum mean discrepancy (MMD) and the adversarial training. In addition, knowledge distillation is also used to perform model-level knowledge transfer.

\textbf{MMD based methods} \cite{DBLP:journals/corr/TzengHZSD14} construct a reproducing kernel Hilbert space (RKHS) $\mathcal{H}_{\kappa}$ with the kernel $\kappa$, and optimize the $\mathcal{H}$-divergence by minimizing the MMD distance $d_{\text{MMD}}^{\kappa}(\sD_{S},\sD_{T})$ on $\mathcal{H}_{\kappa}$. Using the kernel trick, MMD can be computed as 
\begin{equation}
    \begin{split}
        &d_{\text{MMD}}^{\kappa}(\sD_{S},\sD_{T})=-2\E_{\rmX_{S},\rmX_{T}\sim \sD_{S},\sD_{T}}\kappa(\rmX_{S},\rmX_{T})+\\
        &\E_{\rmX_{S},\rmX'_{S} \sim \sD_{S}}\kappa(\rmX_{S},\rmX'_{S})+\E_{\rmX_{T},\rmX'_{T}\sim\sD_{T}}\kappa(\rmX_{T},\rmX'_{T})
    \end{split}
    \label{eq:MMD}
\end{equation}
Recent works propose the variations of MMD, e.g., multi-kernel MMD \citep{DBLP:conf/icml/LongC0J15}, class-weighted MMD\citep{DBLP:conf/cvpr/YanDLWXZ17} and domain-crossing MMD \citep{DBLP:conf/iccv/PengBXHSW19}. However, all these methods require the pairwise calculation of the data from source and target domains, which is not allowed under the decentralization constraints.

\textbf{The adversarial training strategy} \cite{DBLP:conf/cvpr/SaitoWUH18,DBLP:conf/iclr/0002ZWCMG18} apply adversarial training in feature space to optimize $\mathcal{H}$-divergence. It is proved that with the adversarial training strategy, the UMDA model can work under the privacy-preserving policy \citep{DBLP:conf/iclr/PengHZS20}. However, the adversarial training requires each source domain to exchange and update model parameters
with the target domain after every single batch, which consumes huge communication resources.

\textbf{Knowledge distillation in domain adaptation.} Self-supervised learning has many applications \cite{zhang2021cause,chen2021electrocardio} in the label-lacking scenarios.
Knowledge distillation (KD) \cite{DBLP:journals/corr/HintonVD15,DBLP:conf/aaai/ChenMWF020} is an efficient self-supervised method to transfer knowledge between different models. Recent works \cite{DBLP:conf/icassp/MengLGJ18,DBLP:journals/corr/abs-2003-07325} extend the knowledge distillation into domain adaptation with a teacher-student training strategy: training multiple teacher models on source domains and ensembling them on target domain to train a student model. This strategy outperforms other UMDA method in practice. However, due to the
irrelevant and malicious source domains, the conventional
KD strategies may fail to obtain proper knowledge.
\subsection{Federated Learning}
Federated learning \citep{DBLP:journals/corr/KonecnyMYRSB16} is a distributed machine learning approach, it can train a global model by aggregating the updates of local models from multiple decentralized datasets. Recent works \cite{DBLP:conf/aistats/McMahanMRHA17} find a trade-off between model performance and communication efficiency, that is, to make the global model achieve better performance, we need to conduct more communication rounds, which raises the communication costs. Besides, the frequent communication will also cause privacy leakage \cite{DBLP:conf/infocom/WangSZSWQ19}, making the training process insecure.

\textbf{Federated domain adaptation}. There are few works discussing the decentralized UMDA methods. FADA \citep{DBLP:conf/iclr/PengHZS20} first raises the concept of federated domain adaptation. It applies the adversarial training to optimize the $\mathcal{H}$-divergence without accessing data. However, FADA consumes high communication costs and is vulnerable to the privacy leakage attack. Model Adaptation \cite{DBLP:conf/cvpr/LiJ0WW20} and SHOT \cite{DBLP:conf/icml/LiangHF20} provide source-free methods to solve the single
source decentralized domain adaptation. However, they are vulnerable to the negative transfer in multi-source situations.

\section{KD3A: Decentralized Domain Adaptation via Knowledge Distillation}
\textbf{Preliminary.} Let $\sD_{S}$ and $\sD_{T}$ denote the source domain and target domain. In UMDA, we have $K$ \textit{source} domains $\{\sD^{k}_{S}\}_{k=1}^{K}$ where each domain contains $N_{k}$ labeled examples as $\sD^{k}_{S} := \{(\rmX^{k}_{i},\rvy^{k}_{i})\}_{i=1}^{N_k}$ and a \textit{target} domain $\sD_{T}$ with $N_{T}$ unlabeled examples as $\sD_{T}:=\{\rmX^{T}_{i}\}_{i=1}^{N_{T}}$. The goal of UMDA is to learn a model $h$ which can minimize the task risk $\epsilon_{\sD_{T}}$ in $\sD_{T}$, i.e. $\epsilon_{\sD_{T}}(h)=\Pr_{(\rmX,\rvy)\sim \sD_{T}}[h(\rmX)\neq \rvy]$. Without loss of generality, we consider $C$-way classification task and assume the target domain shares the same tasks with the source domains. In a common UMDA, we combine $K$ source domains with different domain weights as $\bm{\alpha}$, and perform domain adaptation by minimizing the following generalization bound \citep{DBLP:journals/ml/Ben-DavidBCKPV10,DBLP:conf/iclr/0002ZWCMG18} with the multiple source domains as:

\textbf{Theorem 1} \textit{Let $\mathcal{H}$ be the model space, $\{\epsilon_{\sD_{S}^{k}}(h)\}_{k=1}^{K}$ and $\epsilon_{\sD_{T}}(h)$ be the task risks of source domains $\{\sD_{S}^{k}\}_{k=1}^{K}$ and the target domain $\sD_{T}$, and $\bm{\alpha} \in \mathcal{R}^{K}_{+},\sum_{k=1}^{K}\bm{\alpha}_{k}=1$ be the domain weights. Then for all $h \in \mathcal{H}$ we have:}
\begin{equation}
\epsilon_{\sD_{T}}(h)\leq\sum_{k=1}^{K}\bm{\alpha}_{k}\left(\epsilon_{\sD_{S}^{k}}(h)+\frac{1}{2}d_{\mathcal{H}\Delta \mathcal{H}}(\sD_{S}^{k},\sD_{T})\right)+\lambda_0
\label{eq:original-bound}
\end{equation}
where $\lambda_0$ is a constant according to the task risk of the optimal model on the source domains and target domain. 

\textbf{Problem formulation for decentralized scenarios.} In decentralized UMDA, the data from $K$ source domains are stored locally and are not available. The accessible information in each communication round includes: (1) The size of the training sets $\{N_{S}^{k}\}_{k=1}^{K}$ on source domains and the parameters of $K$ models $\{h_{S}^{k}\}_{k=1}^{K}$ trained on these source domains. (2) The target domain data containing $N_{T}$ unlabeled examples as $\sD_{T}:=\{\rmX_{i}^{T}\}_{i=1}^{N_T}$. In KD3A, we apply knowledge distillation to perform domain adaptation without accessing the data.

\subsection{Extending Source Domains With Consensus Knowledge}
Knowledge distillation can perform knowledge transfer through different models. Suppose we have $K$ fully-trained models from $K$ source domains denoted by $\{h^{k}_{S}\}_{k=1}^K$. we use $q^{k}_{S}(\rmX)$ to denote the confidence for each class and use the class with the maximum confidence as label, i.e. $h^{k}_{S}(\rmX)=\arg_{c}\max [q^{k}_{S}(\rmX)]_{c}$. As shown in Figure \ref{fig:kdkv}(a), the knowledge distillation in UMDA consists of two steps. First, for each target domain data $\rmX_{i}^{T}$, we obtain the inferences of the source domain models. Then, we use the ensemble method to get the consensus knowledge of the source models, e.g., $\rvp_{i}=\frac{1}{K}\sum_{k=1}^{K}q^{k}_{S}(\rmX_{i}^{T})$. In order to utilize the consensus knowledge for domain adaptation, we define an extended source domain $\sD^{K+1}_{S}$ with the consensus knowledge $\rvp_{i}$ for each target domain data $\rmX_{i}^{T}$ as
\begin{equation*}
\sD^{K+1}_{S}=\{(\rmX^{T}_{i},\rvp_{i})\}_{i=1}^{N_{T}}
\end{equation*}
We also define the related task risk for $\sD^{K+1}_{S}$ as 
\begin{equation*}
    \epsilon_{\sD^{K+1}_{S}}(h)=\Pr_{(\rmX,\rvp)\sim \sD^{K+1}_{S}}[h(\rmX)\neq\arg_{c}\max\rvp_{c}].
\end{equation*}
With this new source domain, we can train the source model $h_{S}^{K+1}$ through the knowledge distillation loss as
\begin{equation}
    L^{\text{kd}}(\rmX_{i}^{T},q_{S}^{K+1})=\KL(\rvp_{i} \Vert q_{S}^{K+1}(\rmX_{i}^{T})).
    \label{eq:kdloss}
\end{equation}
In decentralized UMDA, we get the target model as the aggregation of the source models, i.e. $h_{T}:=\sum_{k=1}^{K+1}\bm{\alpha}_{k}h_{S}^{k}$. A common question is, how does the new model $h_{S}^{K+1}$ improve the UMDA performance? It is easy to find that minimizing KD loss (\ref{eq:kdloss}) leads to the optimization of $\epsilon_{\sD^{K+1}_{S}}(h)$ (proof in Appendix A). With this insight, we can derive the generalization bound for knowledge distillation as follows (proof in Appendix B):
\begin{figure}[t]
\centering
\subfigure[Knowledge distillation process in UMDA.]{
    \includegraphics[width=3in]{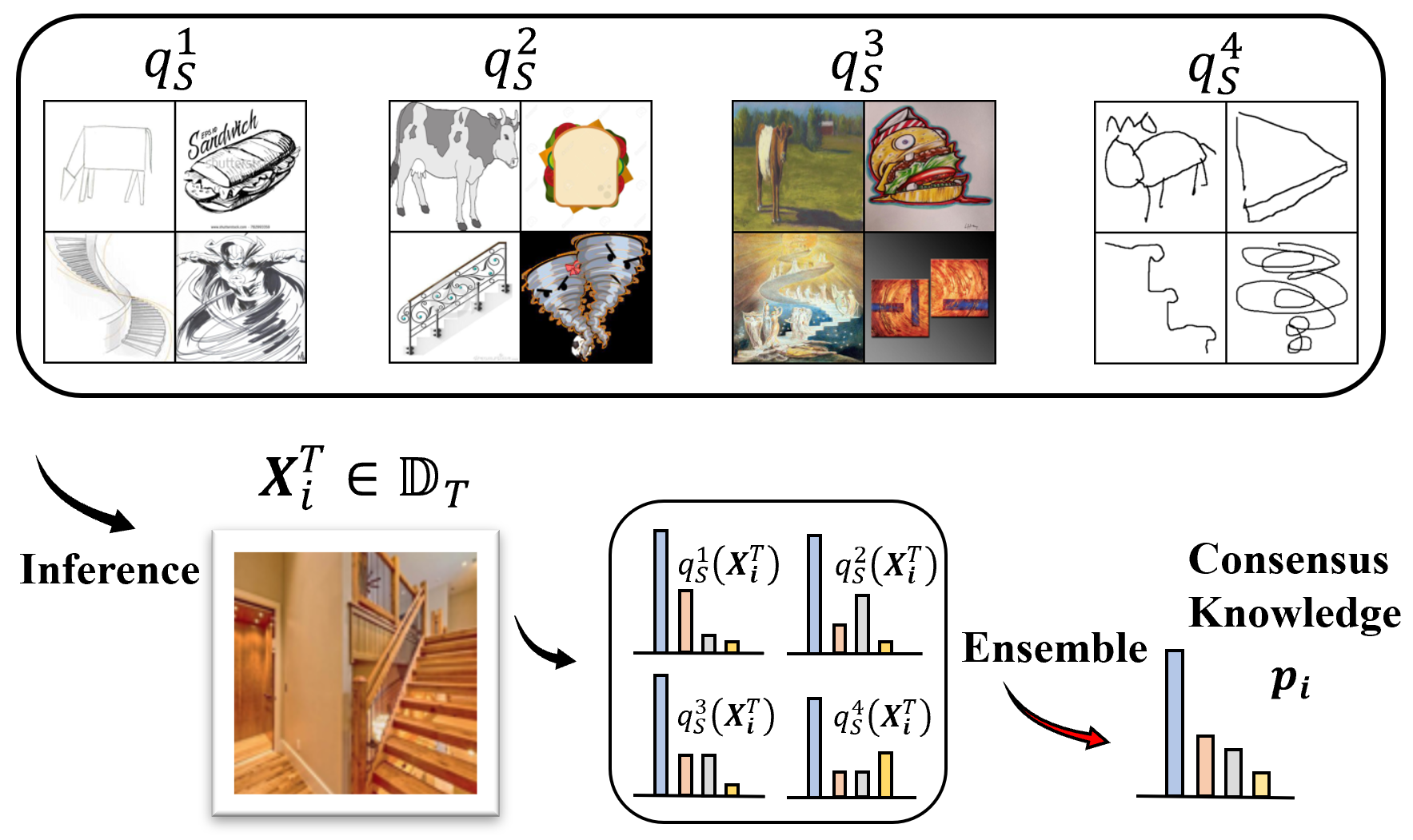}
}
\quad    
\subfigure[Knowledge vote ensemble.]{
\includegraphics[width=3in]{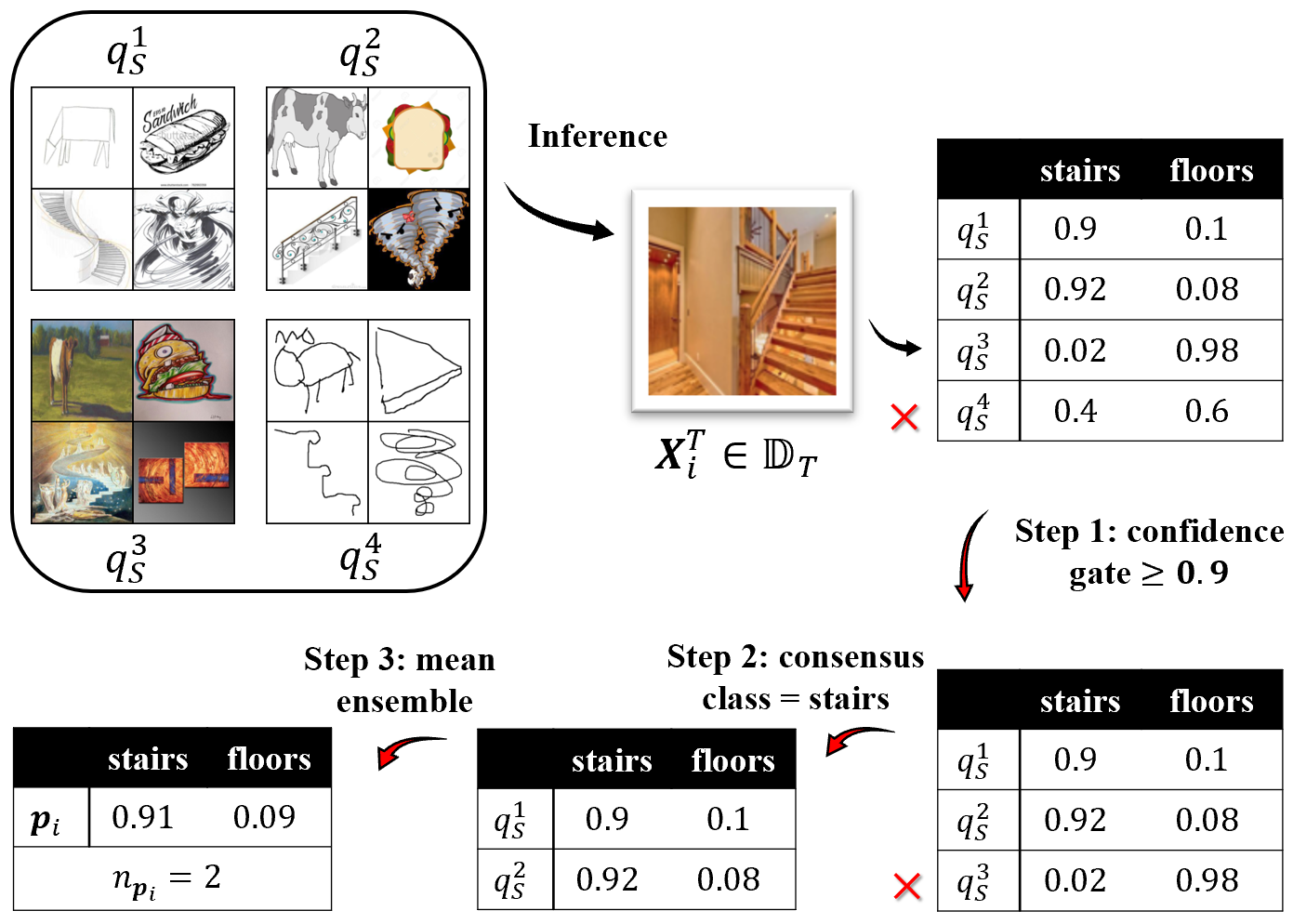}
}
\caption{(a) Knowledge distillation in UMDA consists of two steps: obtaining the inferences from source domain models and performing knowledge ensemble to get the consensus knowledge. (b) Our knowledge vote extracts strong consensus knowledge with 3 steps: confidence gate, consensus class vote and mean ensemble. `\textcolor{red}{$\bm{\times}$}' means the eliminated model in each step.}
\label{fig:kdkv}
\end{figure}

\textbf{Proposition 1} (\textit{The generalization bound for knowledge distillation). Let $\mathcal{H}$ be the model space and $\epsilon_{\sD^{K+1}_{S}}(h)$ be the task risk of the new source domain $\sD^{K+1}_{S}$ based on knowledge distillation. Then for all $h_{T}
\in \mathcal{H}$, we have:}
\begin{equation}
    \begin{split}
        \epsilon_{\sD_{T}}(h_{T})\leq \epsilon_{\sD^{K+1}_{S}}(h_{T})+\frac{1}{2}d_{\mathcal{H}\Delta \mathcal{H}}(\sD_{S}^{K+1},\sD_{T})\\
        +\min\{\lambda_1,
        \sup_{h\in \mathcal{H}}\vert\epsilon_{\sD^{K+1}_{S}}(h)-\epsilon_{\sD_{T}}(h)\vert\}
        \label{eq:kb}    
    \end{split}
\end{equation}
where $\lambda_1$ is a constant for the task risk of the optimal model.

\subsection{Knowledge Vote: Producing Good Consensus}
Proposition 1 shows the new source domain $\sD^{K+1}_{S}$ will improve the generalization bound if the consensus knowledge is good enough to represent the ground-truth label, i.e. $\sup_{h\in \mathcal{H}}\vert\epsilon_{\sD_{T}}(h)-\epsilon_{\sD^{K+1}_{S}}(h)\vert\leq \lambda_1$. However, due to the irrelevant and malicious source domains, the conventional ensemble strategies (e.g., maximum and mean ensemble) may fail to obtain proper consensus. Therefore, we propose the \textit{Knowledge Vote} to provide high-quality consensus. 

The main idea of knowledge vote is that if a certain consensus knowledge is supported by more source domains with high confidence (e.g., $>0.9$), then it will be more likely to be the true label. As shown in Figure \ref{fig:kdkv}(b), it takes three steps to perform \textit{Knowledge Vote}:

\begin{enumerate}
    \item \textbf{Confidence gate}. For each $ \rmX_{i}^{T}\in \sD_{T}$, we firstly use a high-level confidence gate to filter the predictions $\{q^{k}_{S}(\rmX_{i}^{T})\}_{k=1}^K$ of teacher models and eliminate the unconfident models.
    \item \textbf{Consensus class vote}. For the models remained, the predictions are added up to find the consensus class which has the maximum value. Then we drop the models that are inconsistent with the consensus class.
    \item \textbf{Mean ensemble}. After the class vote, we obtain a set of models that all support the consensus class. Finally, we get the consensus knowledge $\rvp_{i}$ by conducting the mean ensemble on these supporting models. We also record the number of domains that support $\rvp_{i}$, denoted by $n_{\rvp_{i}}$. For those $\rmX^{T}$ with all teacher models eliminated by the confidence gate, we simply use the mean ensemble to get $\rvp$ and assign a relatively low weight to them as $n_{\rvp}=0.001$. 
\end{enumerate}
After \textit{Knowledge Vote}, we obtain the new source domain $\sD^{K+1}_{S}=\{(\rmX^{T}_{i},\rvp_{i},n_{\rvp_{i}})\}_{i=1}^{N_T}$. We use the $n_{\rvp_{i}}$ to re-weight the knowledge distillation loss as  
\begin{equation}
    L^{\text{kv}}(\rmX^{T}_{i},q)=n_{\rvp_{i}}\cdot \KL(\rvp_{i}\Vert q(\rmX^{T}_{i})) 
    \label{eq:kvloss}
\end{equation}

Compared with other ensemble strategies, our \textit{Knowledge Vote} makes model learn high-quality consensus knowledge since we assign high weights to those items with high confidence and many support domains. 
\subsection{Consensus Focus: Against Negative Transfer}
Domain weights $\bm{\alpha}$ determine the contribution of each source domain. \citet{DBLP:journals/ml/Ben-DavidBCKPV10} proves the optimal $\bm{\alpha}$ should be proportional to the amount of data when all source domains are 
equally important. However, this condition is hard to satisfy in KD3A since some source domains are usually very different from the target domain, or even malicious domains with corrupted labels. These bad domains lead to negative transfer. One common solution \citep{DBLP:conf/aaai/ZhaoWZGLS0HCK20} is to re-weight each source domain with the $\mathcal{H}$-divergence as 
\begin{equation}
    \bm{\alpha}_{k}=N_{k} e^{-d_{\mathcal{H}}(\sD_{S}^{k},\sD_{T})}/\sum_{k}N_{k} e^{-d_{\mathcal{H}}(\sD_{S}^{k},\sD_{T})}.
    \label{eq:H-reweight}
\end{equation}
However, calculating $\mathcal{H}$-divergence requires to access the source domain data. Besides, $\mathcal{H}$-divergence only measures the domain similarity on the input space, which does not utilize the label information and fails to identify the malicious domain. Reasonably, we propose \textit{Consensus Focus} to identify those irrelevant and malicious domains. As mentioned in \textit{Knowledge Vote}, the UMDA performance is related to the quality of consensus knowledge. With this motivation, the main idea of \textit{Consensus Focus} is to assign high weights to those domains which provide high-quality consensus and penalize those domains which provide bad consensus. To perform \textit{Consensus Focus}, we first derive the definition of consensus quality and then calculate the contribution to the consensus quality for each source domain.

\textbf{The definition of consensus quality.} Suppose we have a set of source domains denoted by $\mathcal{S}=\{\sD_{S}^{k}\}_{k=1}^{K}$. For each coalition of source domains $\mathcal{S}',\mathcal{S}'\subseteq \mathcal{S}$, we want to estimate the quality of the knowledge consensus obtained from $\mathcal{S}'$. Generally speaking, if one consensus class is supported by more source domains with higher confidence, then it will be more likely to represent the true label, which means the consensus quality gets better. Therefore, for each $\rmX_{i}^{T}\in \sD_{T}$ with the consensus knowledge $(\rvp_{i}(\mathcal{S}'),n_{\rvp_{i}}(\mathcal{S}'))$ obtained from $\mathcal{S}'$, We define the related consensus quality as $n_{\rvp_i}(\mathcal{S}') \cdot \max \rvp_{i}(\mathcal{S}')$ and the total consensus quality $Q$ is
\begin{equation}
    Q(\mathcal{S}')=\sum_{\rmX_{i}^{T}\in \sD_{T}}n_{\rvp_i}(\mathcal{S}') \cdot \max \rvp_{i}(\mathcal{S}')
    \label{eq:cq}
\end{equation}
With the consensus quality defined in (\ref{eq:cq}), we derive the consensus focus (CF) value to quantify the contribution of each source domain as 
\begin{equation}
    \text{CF}(\sD_{S}^{k})=Q(\mathcal{S})-Q(\mathcal{S}\setminus\{\sD_{S}^{k}\})
\end{equation}
$\text{CF}(\sD_{S}^{k})$ describes the marginal contribution of the single source domain $\sD_{S}^{k}$ to the consensus quality of all source domains $\mathcal{S}$. If one source domain is a bad domain, then removing it will not decrease the total quality $Q$, which leads to a low consensus focus value. With the CF value, we can assign proper weights to different source domains. Since we introduce a new source domain $\sD_{S}^{K+1}$ in \textit{Knowledge Vote}, we compute the domain weights with two steps. First, we obtain $\bm{\alpha}_{K+1}=N_{T}/(\sum_{k=1}^{K}N_{k}+N_{T})$ for $\sD_{S}^{K+1}$ based on the amount of data. Then we use the CF value to re-weight each original source domain as
\begin{equation}
    \bm{\alpha}_{k}^{\text{CF}} = (1-\bm{\alpha}_{K+1})\cdot\frac{N_{k}\cdot \text{CF}(\sD_{S}^{k})}{\sum_{k=1}^{K} N_{k}\cdot \text{CF}(\sD_{S}^{k})}
\end{equation}

Compared with the re-weighting strategy in (\ref{eq:H-reweight}), our \textit{Consensus Focus} has two advantages. First, the calculation of $\bm{\alpha}^{\text{CF}}$ does not need to access the original data. Second, $\bm{\alpha}^{\text{CF}}$ obtained through \textit{Consensus Focus} is based on the quality of consensus, which utilize both data and label information and can identify malicious domains.

\subsection{BatchNorm MMD: Decentralized Optimization Strategy of $\mathcal{H}-$divergence}
To get a better UMDA performance, we need to minimize the $\mathcal{H}$-divergence between source domains and target domain, where the kernel-based MMD distance is widely used. Existing works \cite{DBLP:conf/icml/LongC0J15,DBLP:conf/iccv/PengBXHSW19} use the feature $\bm{\pi}$ extracted by the fully-connected (fc) layers to build kernel as $\kappa(\rmX^{S},\rmX^{T})=\langle\bm{\pi}^{S},\bm{\pi}^{T}\rangle$ and the related optimization target is
\begin{equation}
        \min_{h\in\mathcal{H}}\sum_{k=1}^{K+1}\bm{\alpha}_{k}d_{\text{MMD}}^{\kappa}(\sD_{S}^{k},\sD_{T})
    \label{eq:MMD-target}
\end{equation}
However, these methods is not applicable in decentralized UMDA since the source domain data is unavailable. Besides, only using the high-level features from fc-layers may lose the detailed 2-D information. Therefore, we propose the \textit{BatchNorm MMD}, which utilizes the mean and variance parameters in each BatchNorm layer to optimize the $\mathcal{H}-$divergence without accessing data.

BatchNorm (BN) \citep{DBLP:conf/icml/IoffeS15} is a widely-used normalization technique. For the feature $\bm{\pi}$, BatchNorm is expressed as $\text{BN}(\bm{\pi}) = \gamma\cdot \frac{\bm{\pi}-\E(\bm{\pi})}{\sqrt{\Var(\bm{\pi})}}+\beta$, where $(\E(\bm{\pi}),\Var(\bm{\pi}))$ are estimated in training process\footnote{Implemented with \textit{running-mean} and \textit{running-var} in Pytorch.}. Supposing the model contains $L$ BatchNorm layers, we consider the quadratic kernel for the feature $\bm{\pi}_{l}$ of the $l$-th BN-layer, i.e. $  \kappa(\rmX^{S},\rmX^{T}) =\left( \langle\bm{\pi}_{l}^{S},\bm{\pi}_{l}^{T} \rangle+\frac{1}{2}\right)^2$. The MMD distance based on this kernel is 
\begin{equation}
\begin{split}
    d_{\text{MMD}}^{\kappa}(\sD_{S}^{k},\sD_{T})=&\Vert\E(\bm{\pi}_{l}^{k})-\E(\bm{\pi}_{l}^{T})\Vert_2^2\\
    +&\Vert\E[\bm{\pi}_{l}^{k}]^2-\E[\bm{\pi}_{l}^{T}]^2\Vert_2^2
\end{split}
    \label{eq:BN-MMD}
\end{equation}

Compared with other works using the quadratic kernel \cite{DBLP:conf/iccv/PengBXHSW19}, we can obtain all required parameters in (\ref{eq:BN-MMD}) through the parameters $(\E(\bm{\pi}_{l}),\Var(\bm{\pi}_{l}))$ of BN-layers in source domain models without accessing data\footnote{Notice $\E[\bm{\pi}]^2=\Var(\bm{\pi})+[\E(\bm{\pi})]^2$}. Based on this advantage, BatchNorm MMD can perform the decentralized optimization strategy of $\mathcal{H}-$divergence with two steps. First, we obtain  $\{(\E(\bm{\pi}_{l}^{k}),\Var(\bm{\pi}_{l}^{k}))\}_{l=1}^{l}$ from the models on different source domains. Then, for every mini-batch $\rmX^{T}\in\sD_{T}$, we train the model $h_{T}$ to optimize the domain adaptation target (\ref{eq:MMD-target}) with the following loss
\begin{equation}
\begin{split}
    \sum_{l=1}^{L}\sum_{k=1}^{K+1}\bm{\alpha}_{k}
    \big (\Vert\mu(\bm{\pi}_{l}^{T})-\E(\bm{\pi}_{l}^{k})\Vert_2^2
    +\Vert\mu[\bm{\pi}_{l}^{T}]^2-\E[\bm{\pi}_{l}^{k}]^2\Vert_2^2 \big )
\end{split}
\label{eq:bmmd}
\end{equation}
where $(\bm{\pi}_{1}^{T},\ldots,\bm{\pi}_{L}^{T})$ are the features of target model $h_{T}$ from BatchNorm layers corresponding to the input $\rmX^{T}$. In training process, We use the mean value $\mu$ of every mini-batch to estimate the expectation $\E$. In addition, optimizing the loss (\ref{eq:bmmd}) requires traversing all Batchnorm layers, which is time-consuming. Therefore, we propose a computation-efficient optimization strategy in Appendix E.
\begin{algorithm}[t] 
\caption{KD3A training process with epoch t.} 
\label{alg:KD3A} 
\begin{algorithmic}[1] 
\REQUIRE ~~\\ 
Source domains $\mathcal{S}=\{\sD_{S}^{k}\}_{k=1}^{K}$. Target domain $\sD_{T}$;\\
Target model $h_{T}^{(t-1)}$ with parameters $\bm{\Theta}^{(t-1)}$;\\
Confidence gate $g^{(t)}$;\\
\ENSURE ~~\\ 
Target model $h_{T}^{(t)}$ with parameters $\bm{\Theta}^{(t)}$.\\
\STATE // Locally training on source domains:
\FOR{$\sD_{S}^{k}$ in $\mathcal{S}$} 
    \STATE Model initialize: $(h^{k}_{S},\bm{\Theta}^{k}_{S}) \leftarrow (h^{(t-1)},\bm{\Theta}^{(t-1)})$.
    \STATE Train $h^{k}_{S}$ with classification loss on $\sD_{S}^{k}$.
\ENDFOR
\STATE Upload $\{(h^{k}_{S},\bm{\Theta}^{k}_{S})\}_{k=1}^{K}$ to the target domain.
\STATE // \textit{Knowledge Vote:}
\STATE $\sD_{S}^{K+1} \leftarrow\text{KnowledgeVote}(\sD_{T},g^{(t)},\{h^{k}_{S}\}_{k=1}^{K})$.
\STATE Train $h^{K+1}_{S}$ with $L^{\text{kv}}$ loss (\ref{eq:kvloss}) on $\sD_{S}^{K+1}$.
\STATE // \textit{Consensus Focus:}
\STATE $\bm{\alpha}^{\text{CF}}\leftarrow\text{ConsensusFocus}(\sD_{T},\{h^{k}_{S}\}_{k=1}^{K},\{N_{k}\}_{k=1}^{K})$.
\STATE // \textit{Model Aggregation:}
\STATE $\bm{\Theta}^{(t)}\leftarrow \sum_{k=1}^{K+1}\bm{\alpha}^{\text{CF}}_{k}\cdot \bm{\Theta}^{k}_{S}$.
\STATE // \textit{BatchNorm MMD:}
\STATE Obtain $\{\E[\bm{\pi}_{l}^{k}]^{i},i=1,2\}_{l,k=1}^{L,K+1}$ from $\{(h^{k}_{S},\bm{\Theta}^{k}_{S})\}_{k=1}^{K+1}$
\STATE Train $h_{T}^{(t)}$ with BatchNorm MMD on $\sD_{T}$.
\STATE Return $(h_{T}^{(t)},\bm{\Theta}^{(t)})$.
\end{algorithmic}
\end{algorithm}
\subsection{The Algorithm of KD3A}
In the above sections, we have proposed three essential components that work well in KD3A, and the complete algorithm of KD3A can be obtained by using these components in tandem: First, we obtain an extra source domain $\sD^{K+1}_{S}$ and train the source model $h^{K+1}_{S}$ through  \textit{Knowledge Vote}. Then, we get the target model by aggregating $K+1$ source models through  \textit{Consensus Focus}, i.e. $h_{T}:=\sum_{k=1}^{K+1}\bm{\alpha}_{k}h_{S}^{k}$. Finally, we minimize the $\mathcal{H}-$divergence of the target model through  \textit{Batchnorm MMD}. The decentralized training process of KD3A is shown in Algorithm \ref{alg:KD3A}. Confidence gate is the only hyper-parameter in KD3A, and should be treated carefully. If the confidence gate is too large, almost all data in target domain would be eliminated and the knowledge vote loss would not work. If too small, then the consensus quality would be reduced. Therefore, we gradually increase it from low (e.g., $0.8$) to high (e.g., $0.95$) in training. 
\section{Generalization Bound For KD3A}
We further derive the generalization bound for KD3A by combining the original bound (\ref{eq:original-bound}) and the knowledge distillation bound (\ref{eq:kb}). The related generalization bound is:

\textbf{Theorem 2} \textit{(The decentralized generalization bound for KD3A). Let $h_{T}$ be the target model of KD3A, $\{\sD^{k}_{S}\}_{k=1}^{K+1}$ be the extended source domains through Knowledge Vote and $\bm{\alpha}^{\text{CF}} \in \mathcal{R}^{K+1}_{+},\sum_{k=1}^{K+1}\bm{\alpha}^{\text{CF}} _{k}=1$ be the domain weights through Consensus Focus. Then we have:}
\begin{equation}
 \epsilon_{\sD_{T}}(h_{T})\leq \sum_{k=1}^{K+1}\bm{\alpha}_{k}^{\text{CF}}\left(\epsilon_{\sD^{k}_{S}}(h_{T})+\frac{1}{2}d_{\mathcal{H}\Delta\mathcal{H}}(\sD^{k}_{S},\sD_{T})\right)+\lambda_2
\label{eq:kvb}
\end{equation}

The generalization performance of KD3A bound (\ref{eq:kvb}) depends on the quality of the consensus knowledge, as the following proposition shows (see Appendix C for proof):

\textbf{Proposition 2} \textit{The KD3A bound (\ref{eq:kvb}) is a tighter bound than the original bound (\ref{eq:original-bound}), if the task risk gap between the knowledge distillation domain $\sD^{K+1}_{S}$ and the target domain $\sD_{T}$ is smaller than the following upper-bound for all source domain $k \in \{1,\cdots,K\}$, that is, $\epsilon_{\sD^{K+1}_{S}}(h)$ should satisfy:}
\begin{equation*}
\centering
\begin{split}
         \sup_{h\in \mathcal{H}}\vert\epsilon_{\sD^{K+1}_{S}}(h)-\epsilon_{\sD_{T}}(h)\vert &\leq 
        \inf_{h\in \mathcal{H}}\vert\epsilon_{\sD^{K+1}_{S}}(h)-\epsilon_{\sD_{S}^{k}}(h)\vert\\+\frac{1}{2}d_{\mathcal{H}\Delta\mathcal{H}}&(\sD_{S}^{k},\sD_{T})+\lambda_{S}^{k}
\end{split}
\end{equation*}

Proposition 2 points out two tighter bound conditions: (1) For those good source domains with small $\mathcal{H}-$divergence and low optimal task risk $\lambda_{S}^{k}$, the model should take their advantages to provide better consensus knowledge, i.e. the task risk $\epsilon_{\sD^{K+1}_{S}}$ gets close enough to $\epsilon_{\sD^{T}}$. (2) For those irrelevant and malicious source domains with high $\mathcal{H}-$divergence and $\lambda$, the model should filter out their knowledge, i.e. the task risk $\epsilon_{\sD^{K+1}_{S}}$ stays away from that for bad domains. 

The KD3A has heuristically achieved the above two conditions through the \textit{Knowledge Vote} and \textit{Consensus Focus}: (1) For good source domains, KD3A provides better consensus knowledge with \textit{Knowledge Vote}, making $\epsilon_{\sD^{K+1}_{S}}$ closer to $\epsilon_{\sD^{T}}$. (2) For bad domains, KD3A filters out their knowledge with \textit{Consensus Focus}, making $\epsilon_{\sD^{K+1}_{S}}$ stay away from that for bad domains. We also conduct sufficient experiments to show our KD3A achieves tighter bound with better performance than other UMDA approaches.
\begin{figure}[t]
\centering
\includegraphics[width=2.95in]{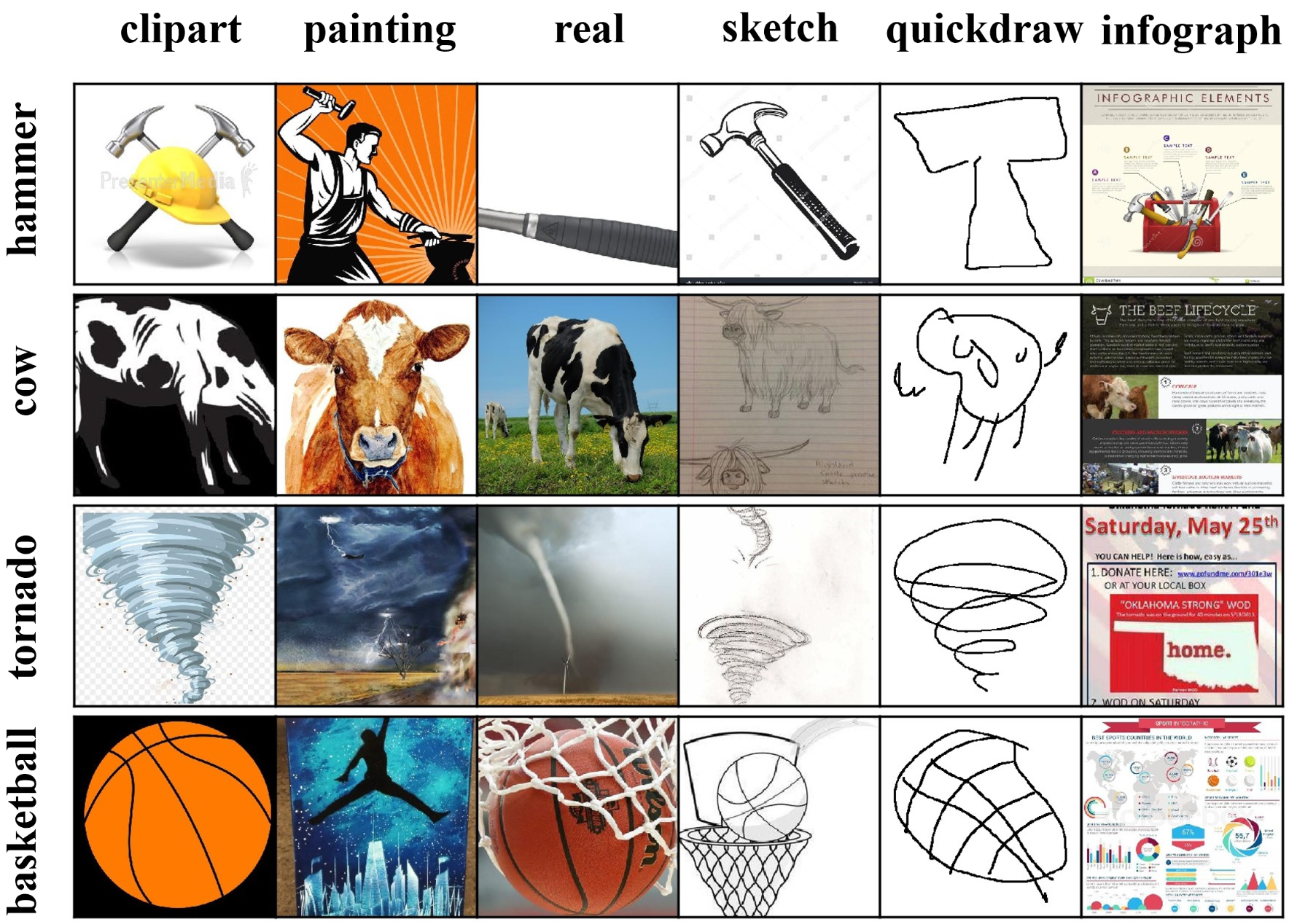}
\caption{The large-scale dataset DomainNet. \textit{Real} is a domain of high quality containing real-world images, while \textit{Quickdraw} is an irrelevant source domain and may cause the negative transfer.}
\label{fig:DomainNet}
\end{figure}
\begin{table*}[t]
\setlength\extrarowheight{4.5pt}
\centering
\begin{tabular}{c|c|cccccc|c}
Standards& Methods & \textit{Clipart} & \textit{Infograph} & \textit{Painting} & \textit{Quickdraw} & \textit{Real} & \textit{Sketch} & Avg \\ 
\hline
\multirow{2}{*}{W/o DA} 
& Oracle &     $69.3_{\pm 0.37}$    &   $34.5_{\pm0.42}$        &     $66.3_{\pm 0.67}$     &   $66.8_{\pm 0.51}$   & $80.1_{\pm 0.59}$      &    $60.7_{\pm 0.48}$    &  $63.0$   \\
\cline{2-9}
& Source-only       &   $52.1_{\pm0.51}$      &    $23.1_{\pm0.28}$       &    $47.7_{\pm0.96}$      &   $13.3_{\pm0.72}$        &   $60.7_{\pm0.32}$   &    $46.5_{\pm0.56}$    &  $40.6$   \\ \hline
\multirow{2}{*}{$\mathcal{H}-$div.}                     
& MDAN    & $60.3_{\pm 0.41}$ &  $25.0_{\pm 0.43}$     &     $50.3_{\pm 0.36}$     &   $8.2_{\pm 1.92}$       &  $61.5_{\pm 0.46}$    &    $51.3_{\pm 0.58}$    &  $42.8$   \\
\cline{2-9}
& $\text{M}^{3}\text{SDA}$ &   $58.6_{\pm0.53}$      &    $\mathbf{26.0_{\pm 0.89}}$       & $52.3_{\pm 0.55}$          &       $6.3_{\pm 0.58}$    &  $62.7_{\pm0.51}$    &  $49.5_{\pm0.76}$      &   $42.6$  \\ 
\hline
\makecell[c]{Knowledge\\ Ensemble}
& DAEL  &    $70.8_{\pm 0.14}$     &    $26.5_{\pm0.13}$       & $57.4_{\pm0.28}$  &  $12.2_{\pm0.7}$         &    $65.0_{\pm 0.23}$   &   $60.6_{\pm0.25}$    &   $48.7$  \\
\hline
\makecell[c]{Source \\ Selection}& CMSS  &  $ 64.2_{\pm 0.18}  $    &  $  28.0_{\pm0.2} $      & $53.6_{\pm0.39}  $       &$16.0_{\pm0.12}  $        &$63.4_{\pm0.21}  $    &  $53.8_{\pm0.35}$    &  $46.5$   \\ 
\hline
Others & DSB$\text{N}^*$       &  $60.3$       &     $22.6$      &    $52.3$      &    $9.1$       &   $62.7$   &   $47.6$     &  $42.4$   \\ 
\hline
\multirow{4}{*}{\makecell[c]{Decentralized\\ UMDA}}     
& SHO$\text{T}^*$ & $61.7$ & $22.2$ & $52.6$ & $12.2$ & $67.7$&$48.6$ & $44.2$\\
\cline{2-9}
   & FAD$\text{A}^*$   &  $59.1$    &     $21.7$      &    $47.9$      &     $8.8$      &   $60.8$   &    $50.4$    &  $41.5$   \\
\cline{2-9}
       & FADA   &  $45.3_{\pm 0.7}$    &     $16.3_{\pm 0.8}$      &    $38.9_{\pm 0.7}$      &     $7.9_{\pm 0.4}$      &   $46.7_{\pm 0.4}$   &    $26.8_{\pm 0.4}$    &  $30.3$   \\
\cline{2-9}
 & KD3A   &   $\mathbf{72.5_{\pm 0.62}}$     & $23.4_{\pm 0.43}$      &    $\mathbf{60.9_{\pm 0.71}}$     &    $\mathbf{16.4_{\pm 0.28}}$        &  $\mathbf{72.7_{\pm 0.55}}$     &   $\mathbf{60.6_{\pm 0.32}}$    &  $\mathbf{51.1}$   \\ 
 \hline
\end{tabular}
\caption{UMDA accuracy $(\%)$ on the DomainNet dataset. Our model KD3A achieves $51.1\%$ accuracy, significantly outperforming all other baselines. Moreover, KD3A achieves the oracle performance on two domains: clipart and sketch. *: The best results recorded in our re-implementation.}
\label{table:dataset}
\end{table*}
\section{Experiments}
\subsection{Domain Adaptation Performance}
We perform experiments on four benchmark datasets: (1) \textbf{Amazon Review} \cite{DBLP:conf/nips/Ben-DavidBCP06}, which is a sentimental analysis dataset including four domains. (2) \textbf{Digit-5} \citep{DBLP:conf/aaai/ZhaoWZGLS0HCK20}, which is a digit classification dataset including five domains. (3) \textbf{Office-Caltech10} \citep{DBLP:conf/cvpr/GongSSG12}, which contains ten object categories from four domains. (4) \textbf{DomainNet} \citep{DBLP:conf/iccv/PengBXHSW19}, which is a recently introduced benchmark for large-scale multi-source domain adaptation with 345 classes and six domains, i.e. \textit{Clipart (clp)}, \textit{Infograph (inf)}, \textit{Painting (pnt)}, \textit{Quickdraw (qdr)}, \textit{Real (rel)} and \textit{Sketch (skt)}, as shown in Figure \ref{fig:DomainNet}. We follow the protocol used in prevailing works, selecting each domain in turn as the target domain and using the rest domains as source domains. Due to space limitations, we mainly present results on DomainNet; more results on Amazon Review, Digit-5 and Office-Caltech10 are provided in Appendix.

\textbf{Baselines.} We conduct extensive comparison experiments with the current best UMDA approaches from 4 categories: (1) $\mathcal{H}$-divergence based methods, i.e. the multi-domain adversarial network (MDAN) \cite{DBLP:conf/iclr/0002ZWCMG18} and moment matching ($\text{M}^3\text{SDA}$) \citep{DBLP:conf/iccv/PengBXHSW19}. (2) Knowledge ensemble based methods, i.e. the domain adaptive ensemble learning (DAEL) \cite{DBLP:journals/corr/abs-2003-07325}. (3) Source selection based methods, i.e. the curriculum manager (CMSS) \cite{DBLP:conf/eccv/YangBLS20}. (4) Decentralized UMDA, i.e. SHOT \citep{DBLP:conf/icml/LiangHF20} and FADA \citep{DBLP:conf/iclr/PengHZS20}. The DSBN proposes a domain-specific BatchNorm, which is similar to Batchnorm MMD, so we also take it into comparison. In addition, We report two baselines without domain adaptation, i.e. oracle and source-only. Oracle directly performs supervised learning on target domains and source-only naively combines source domains to train a single model. 
\begin{figure}[t]
\centering
\includegraphics[width=3in]{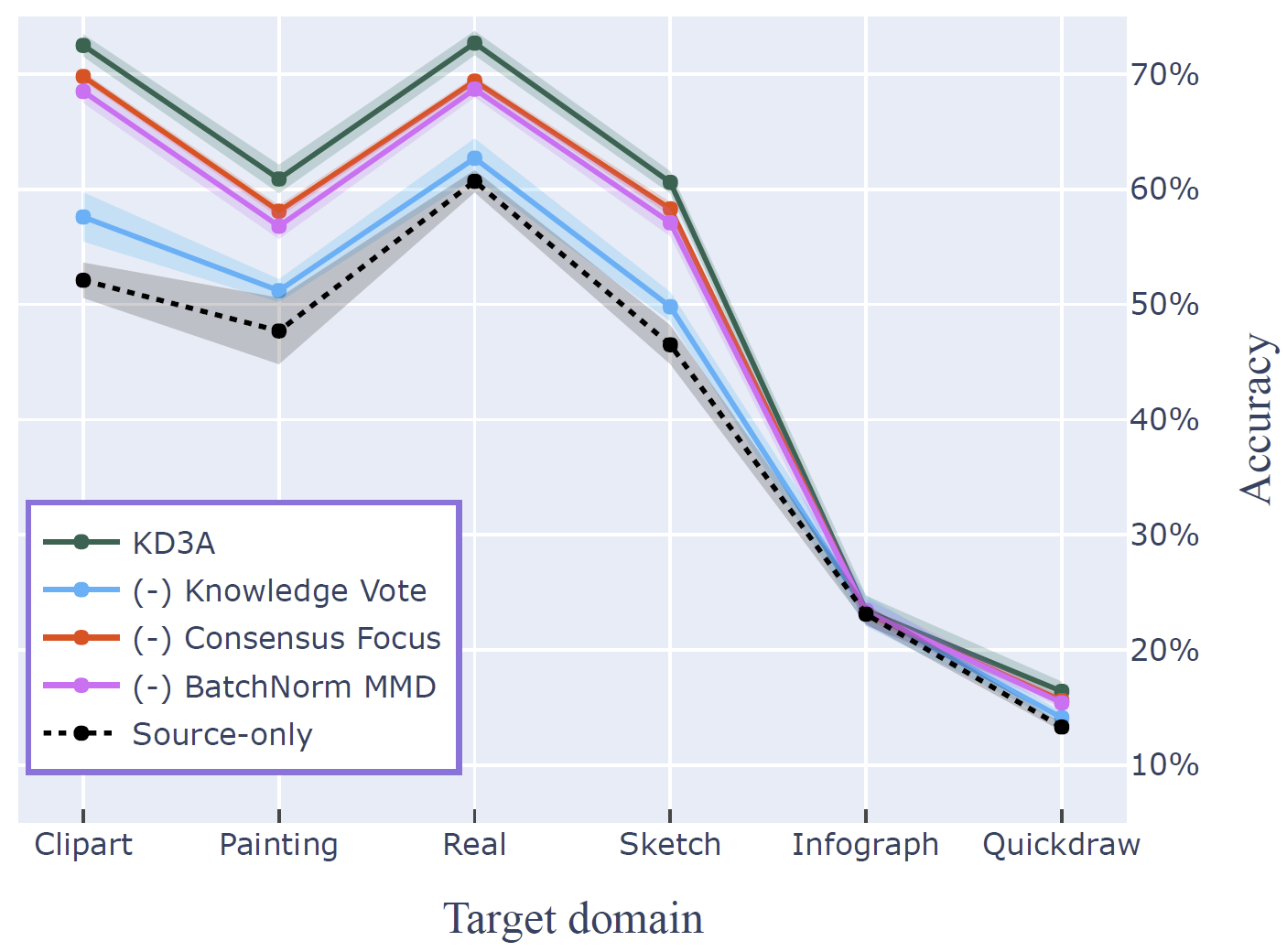}
\caption{The ablation study of KD3A. Results show that \textit{Knowledge Vote}, \textit{Consensus Focus} and \textit{BatchNorm MMD} all contribute to the UMDA performance in all target domains.}
\label{fig:DFDA-ablation}
\end{figure}

\textbf{Implementation details.} Following the settings in previous UMDA works \cite{DBLP:conf/iccv/PengBXHSW19,DBLP:conf/eccv/YangBLS20}, we use a 3-layer MLP as backbone for Amazon Review, a 3-layer CNN for Digit-5 and the ResNet101 pre-trained on ImageNet for Office-Caltech10 and DomainNet. The settings of communication rounds $r$ is important in decentralized training. Since the models on different source domains have different convergence rates, we need to aggregate models $r$ times per epoch. To perform the $r$-round aggregation, we uniformly divide one epoch into $r$ stages and aggregate model after each stage. The KD3A Algorithm \ref{alg:KD3A} is a decentralized training strategy with $r=1$ and we use this setting in all experiments. For model optimization, We use the SGD with 0.9 momentum as the optimizer and take the cosine schedule to decay learning rate from high (i.e. 0.05 for Amazon Review and Digit5, and 0.005 for Office-Caltech10 and DomainNet) to zero. We conduct each experiment five times and report the results with the form $\text{mean}_{\pm \text{std}}$. Since SHOT and DSBN do not report the results on DomainNet, we re-implement them with the official code and report the best testing results.

\textbf{DomainNet.} The results on DomainNet are presented in Table \ref{table:dataset}. In general, our KD3A outperforms all the baselines by a large margin and achieves the oracle performance on \textit{Clipart} and \textit{Sketch}. In addition, compared with the oracle result ($66.8\%$) and the source-only baseline ($13.3\%$), all UMDA methods have failed in \textit{Quickdraw}. Since the \textit{Knowledge Vote} can provide good pseudo-labels for a few good samples and assign low weights to bad samples, KD3A slightly outperforms the source-only baseline. Table \ref{table:dataset} also shows the UMDA performance can benefit from the knowledge ensemble (DAEL) and source domain selection (CMSS). Compared with DAEL, the KD3A provides better consensus knowledge on the high-quality domains such as \textit{Clipart} and \textit{Real}, while it also identifies the bad domains such as \textit{Quickdraw}. CMSS select domains by checking the quality of each data with an independent network. Compared with CMSS, the KD3A does not introduce additional modules and can perform source selection in privacy-preserving scenarios. Moreover, our KD3A outperforms other decentralized models (e.g., SHOT and FADA) through the advantages in knowledge ensemble and source selection. 

\textbf{Ablation study.} To evaluate the contributions of each component, We perform ablation study for KD3A , as shown in Figure \ref{fig:DFDA-ablation}. \textit{Knowledge Vote}, \textit{Consensus Focus} and \textit{Batchnorm MMD} are all able to improve the accuracy, while most contributions are from \textit{Knowledge Vote}, which indicates our KD3A can also perform well on those tasks that cannot use Batchnorm MMD.
\begin{table}[t]
\setlength\extrarowheight{4.5pt}
\begin{tabular}{c|ccc|c}
      &\makecell[c]{$\mathcal{H}$-divergence} & \makecell[c]{Info\\gain}
      & \makecell[c]{Consensus\\ focus} &  \makecell[c]{Domain\\ drop} \\ \hline
IR-qdr & $57.9$ & $57.7$  & $58.1$     & $\mathbf{58.3}$  \\\cline{1-5} 
MA-15  & $50.5$ &$50.5$  & $\mathbf{52.1}$     & \multirow{3}{*}{$50.7$}  \\
MA-30  & $49.8$ &$48.9$  & $\mathbf{51.1}$     &  \\
MA-50  & $47.6$ & $46.3$  &  $50.6$     &  \\
\hline
\end{tabular}
\caption{Average UMDA accuracy ($\%$) with irrelevant and malicious domains. IR-qdr means to use the \textit{Quickdraw} as the irrelevant source domain, while MA-m means to construct a malicious source domain with $m\%$ mislabeled data. With consensus focus, our KD3A is robust to negative transfer.}
\label{table:robust}
\end{table}
\begin{figure}[t]
\centering
\subfigure[IR-qdr.]{
    \includegraphics[height=1.25in]{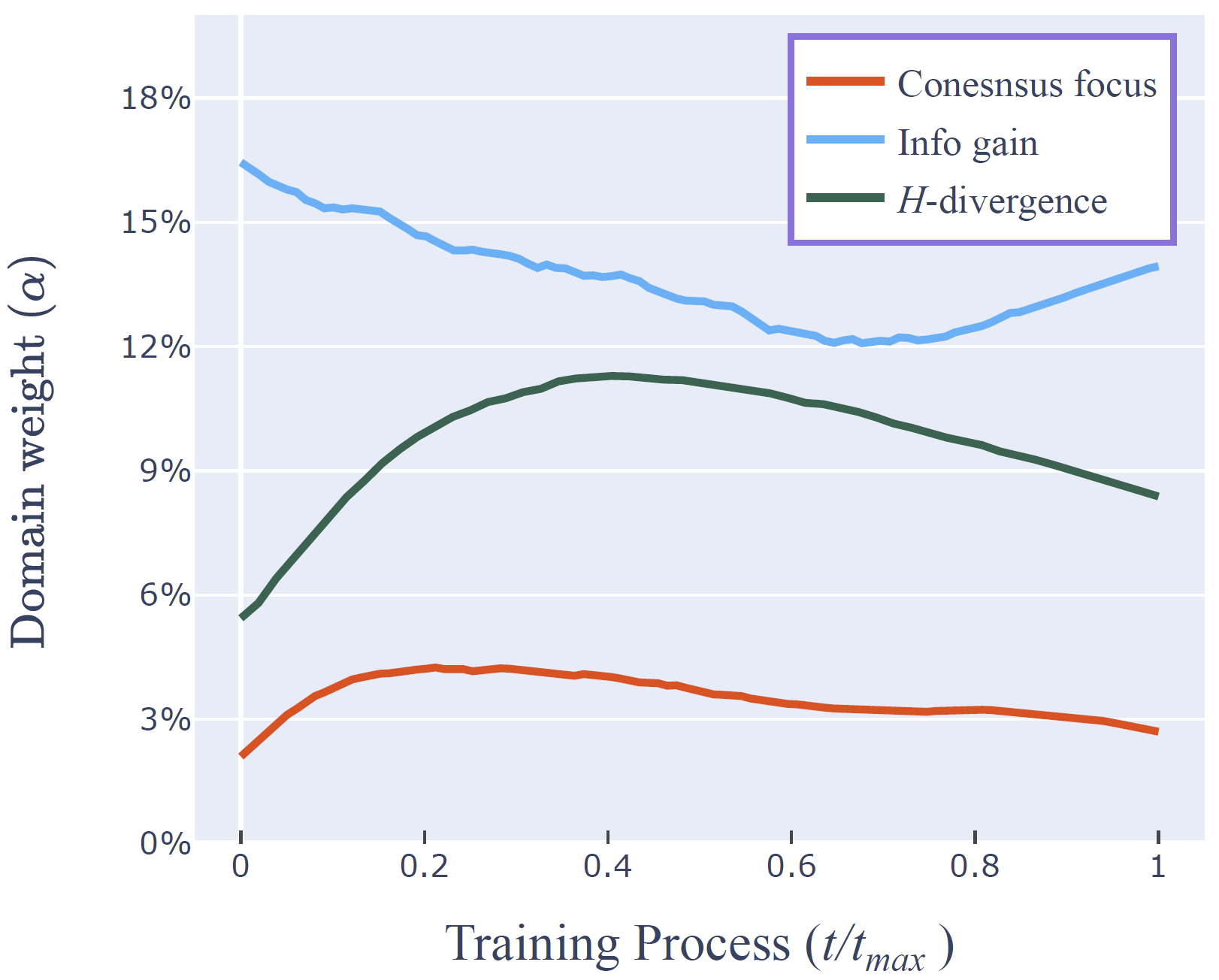}
}
\subfigure[MA-30.]{
\includegraphics[height=1.25in]{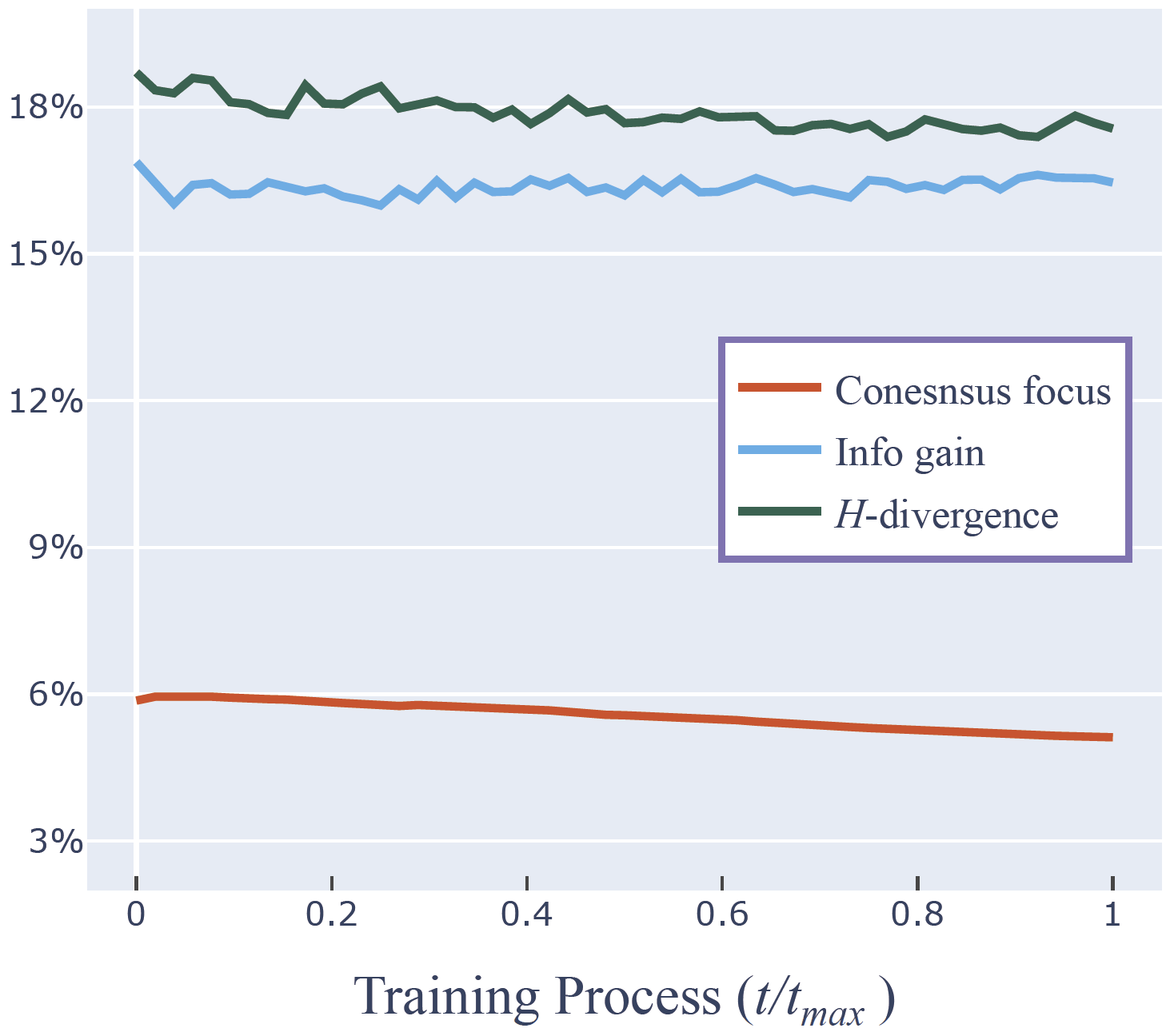}
}
\caption{Weights assigned to the irrelevant and malicious domains in the training process. Our consensus focus can identify these bad domains with the low weights.}
\label{fig:irma}
\end{figure}
\subsection{Robustness To Negative Transfer}
We construct irrelevant and malicious source domains on DomainNet and conduct synthesized experiments to show that with \textit{Consensus Focus}, our KD3A is robust to negative transfer. 

Since \textit{\textit{Quickdraw}} is very different from other domains, and all models perform bad on it, we take \textit{Quickdraw} as the irrelevant domain, denoted by \textbf{IR-qdr}. To construct malicious domains, we perform poisoning attack \citep{DBLP:conf/aistats/BagdasaryanVHES20} on the high-quality domain \textit{Real} with $m\%$ wrong labels, denoted by \textbf{MA-m}. For the irrelevant domain IR-qdr, we select the remaining five domains in turn as target domains and train KD3A with the rest source domains. In training process, we plot the curve of the mean weight $\bm{\alpha}$ assigned to IR-qdr by \textit{Consensus Focus}. We also report the average UMDA accuracy across all target domains. For the malicious domain MA-m, we conduct the same process on the remained four domains except for \textit{Quickdraw}. We report the same experiment results as IR-qdr. 

We consider two advanced weighting strategies 
for comparison: the $\mathcal{H}$-divergence re-weighting in equation (\ref{eq:H-reweight}) and the \textit{Info Gain} in FADA \citep{DBLP:conf/iclr/PengHZS20}. In addition, we also report the average UMDA accuracy of KD3A model with the bad domain dropped. According to the results provided in Table \ref{table:robust} and Figure \ref{fig:irma}, we can get the following insights: (1) For IR-qdr and MA-(30,50), the negative transfer occurs since the domain-drop outperforms the others. (2) The three weighting strategies are robust to the irrelevant domain since they all assign low weights to IR-qdr. (3) \textit{Consensus Focus} outperforms other strategies in malicious domains since it assigns extremely low weights to the bad domain (i.e. $5\%$ for MA-30), while other strategies can not identify the malicious domain. Moreover, our KD3A can use the correct information of less malicious domains (i.e. MA-(15,30)) and achieves better performance than the domain-drop.
\begin{table}[t]
\setlength\extrarowheight{4.5pt}
\begin{tabular}{c|cccccc}
  $r$    &$0.2$ &$0.5$ &$1$ &$2$ &$10$ & $100$ \\ \hline
FADA & $39.2$&$40.3$&$40.5$ & $40.5$  & $40.8$     & $41.5$  \\ \hline
KD3A  & $\mathbf{50.5}$ &$\mathbf{50.9}$  & $\mathbf{51.1}$ & $\mathbf{51.3}$ & $\mathbf{51.3}$     & $\mathbf{52.0}$  \\
\hline
\end{tabular}
\caption{Average UMDA accuracy ($\%$) with different communication rounds $r$ for our KD3A and FADA. KD3A achieves good performance with low communication cost (e.g., $r\leq 1$). }
\label{table:robust-to-cr}
\end{table}
\begin{figure}[t]
\centering
\includegraphics[width=3.2in]{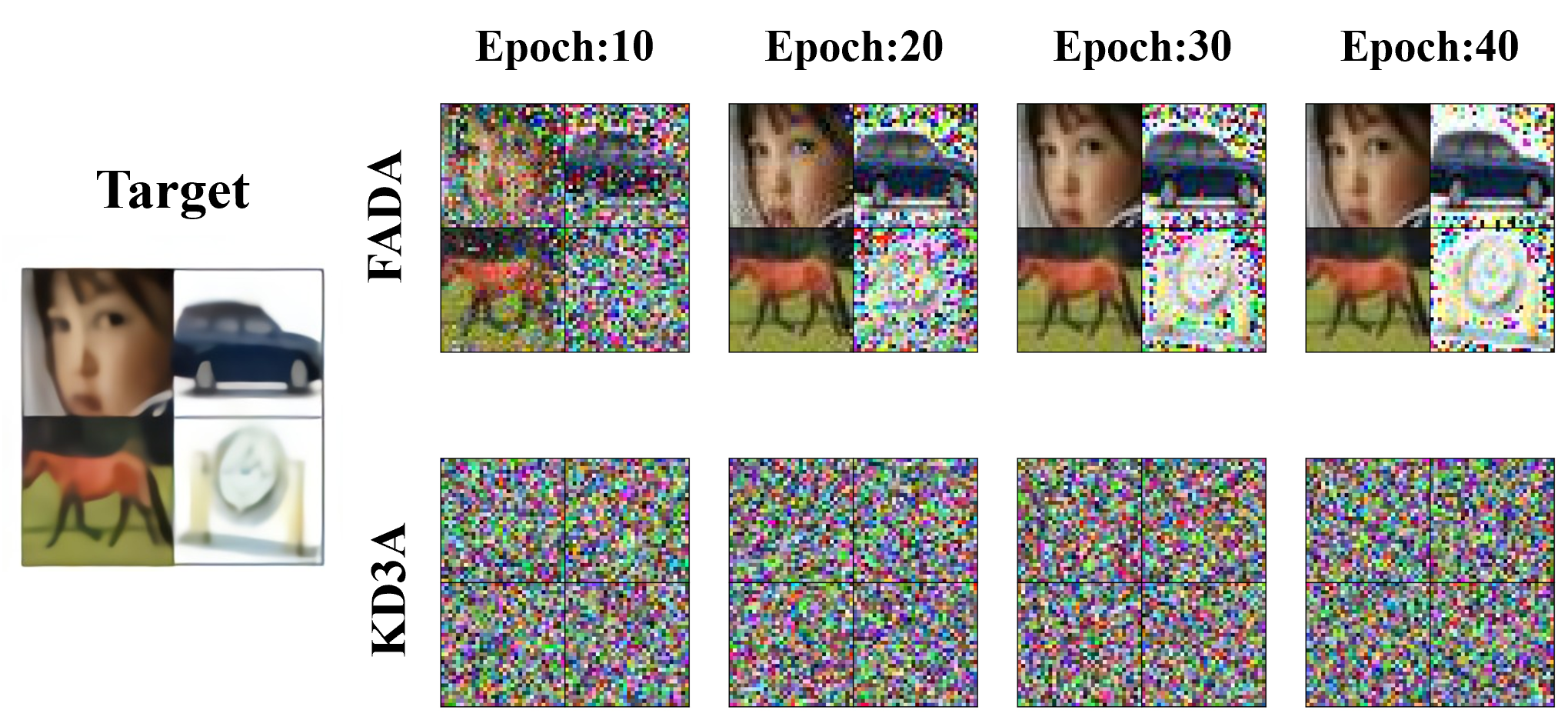}
\caption{The gradient leakage attack \citep{DBLP:conf/nips/ZhuLH19} on decentralized training strategy. KD3A is robust to this attack while FADA causes the privacy leakage.}
\label{fig:gl}
\end{figure}
\subsection{Communication Efficiency And Privacy Security}
To evaluate the communication efficiency, We train the KD3A with different communication rounds $r$ and report the average UMDA accuracy on DomainNet. We take the FADA method as a comparison. The results in Table \ref{table:robust-to-cr} show the following properties: (1) Due to the adversarial training strategy, FADA works under large communication rounds (i.e. $r$ = $100$). (2) Our KD3A works under the low communication cost with $r$ = $1$, leading to a 100 $\times$ communication reduction. (3) KD3A is robust to communication rounds. For example, the accuracy only drops $0.9\%$ when $r$ decreases from $100$ to $1$. Moreover, we consider two extreme cases where we synchronize models every 2 and 5 epochs, i.e. $r$ = $0.5$ and $0.2$. In these cases, FADA performs worse than the source-only baseline while our KD3A can still achieve \text{state-of-the-art} results. 

In decentralized training process, the frequent communication will cause privacy leakage \cite{DBLP:conf/infocom/WangSZSWQ19}, making the training process insecure. To verify the privacy protection capabilities, we perform the advanced gradient leakage attack \citep{DBLP:conf/nips/ZhuLH19} on KD3A and FADA. As shown in Figure \ref{fig:gl}, the source images used in FADA are recovered under the attack, which causes privacy leakage. However, due to the low communication cost, our KD3A is robust to this attack, which demonstrates high privacy security.
\section{Conclusions}
We propose an effective approach KD3A to address the problems in decentralized UMDA. The main idea of KD3A is to perform domain adaptation through the knowledge distillation without accessing the source domain data. Extensive experiments on the large-scale DomainNet demonstrate that our KD3A outperforms other \textit{state-of-the-art} UMDA approaches and is robust to negative transfer. Moreover, KD3A has a great advantage in communication efficiency and is robust to the privacy leakage attack.
\section*{Acknowledgments}
This research has been supported by National Key Research and Development Program (2019YFB1404802) and National Natural Science Foundation of China (U1866602, 61772456).
\newpage
\setcounter{equation}{0}
\section{Appendix}
\subsection{Appendix A}
\textbf{Claim} \textit{For the extended source domain $\sD_{S}^{K+1}=\{(\rmX_{i}^{T},\rvp_{i})\}_{i=1}^{N_{T}}$, training the related source model $h_{S}^{K+1}$ with the knowledge distillation loss $L^{\text{kd}}(\rmX_{i}^{T},q_{S}^{K+1})=\KL(\rvp_{i}\Vert q_{S}^{K+1}(\rmX_{i}^{T}))$ equals to optimizing the task risk $\epsilon_{\sD^{K+1}_{S}}(h)=\Pr_{(\rmX,\rvp)\sim \sD^{K+1}_{S}}[h(\rmX)\neq\arg_{c}\max\rvp_{c}]$.}

\textbf{Proof:}

First, we prove that $\forall c=1,\ldots,C$,
\begin{equation}
    \vert q_{S}^{K+1}(\rmX_{i}^{T}))_{c}-\rvp_{i,c} \vert \leq\sqrt{\frac{1}{2}\KL(\rvp_{i}\Vert q_{S}^{K+1}(\rmX_{i}^{T}))}
\end{equation}
The widely used \textbf{Pinsker's inequality} states that, if $P$ and $Q$ are two probability distributions on a measurable space $(\rmX,\Sigma)$, then 
\[
\delta(P,Q)\leq \sqrt{\frac{1}{2}\KL(P\Vert Q)}
\]
where 
\begin{equation*}
    \begin{split}
        \delta(P,Q) =\sup\{\vert P(\rmA)-Q(\rmA)\vert\vert \rmA \in \Sigma,&\\ \Sigma\text{ is a measurable event.} \}&
    \end{split}
\end{equation*}
In our situation, we choose the event $\rmA$ as the probability of classifying the input $\rmX_{i}^{T}$ into class $c$, and the related probability under $P,Q$ is $\rvp_{i,c}$ and $q_{S}^{K+1}(\rmX_{i}^{T}))_{c}$. With \textbf{Pinsker's inequality}, it is easy to prove $(1)$.  Since the inequality $(1)$ holds for all class $c$, minimizing the knowledge distillation loss will make $q_{S}^{K+1}(\rmX_{i}^{T}))\rightarrow \rvp_i$, that is, $\epsilon_{\sD^{K+1}_{S}}(h)\rightarrow 0$.
\subsection{Appendix B}
\textbf{Proposition 1} (\textit{The generalization bound for knowledge distillation). Let $\mathcal{H}$ be the model space and $\epsilon_{\sD^{K+1}_{S}}(h)$ be the task risk of the new source domain $\sD^{K+1}_{S}$ based on knowledge distillation. Then for all $h_{T}
\in \mathcal{H}$, we have:}
\begin{equation}
    \begin{split}
        \epsilon_{\sD_{T}}(h_{T})\leq \epsilon_{\sD^{K+1}_{S}}(h_{T})+\frac{1}{2}d_{\mathcal{H}\Delta \mathcal{H}}(\sD_{S}^{K+1},\sD_{T})\\
        +\min\{\lambda_1,
        \sup_{h\in \mathcal{H}}\vert\epsilon_{\sD^{K+1}_{S}}(h)-\epsilon_{\sD_{T}}(h)\vert\}
    \end{split}
\end{equation}
where $\lambda_1$ is a constant for the task risk of the optimal model.

\textbf{Proof:}

Following the Theorem 2 in \citet{DBLP:journals/ml/Ben-DavidBCKPV10}, for the source domain $\sD_{S}^{K+1}$ and the target domain $\sD_{T}$, for all $h_{T}\in\mathcal{H}$, we have
\begin{equation}
    \epsilon_{\sD_{T}}(h_{T})\leq \epsilon_{\sD^{K+1}_{S}}(h_{T})+\frac{1}{2}d_{\mathcal{H}\Delta\mathcal{H}}(\sD^{K+1}_{S},\sD_{T})+\lambda_{1}
\end{equation}
where $\lambda_{1}$ is constant of the optimal model on the source domain and the target domain as $\lambda_1=\min_{h\in\mathcal{H}}\epsilon_{\sD^{K+1}_{S}}(h)+ \epsilon_{\sD_{T}}(h)$. 

In addition, the following inequality also holds for all $h_{T}\in\mathcal{H}$:
\begin{equation}
    \epsilon_{\sD_{T}}(h_{T})- \epsilon_{\sD^{K+1}_{S}}(h_{T}) \leq \sup_{h\in \mathcal{H}}\vert\epsilon_{\sD_{T}}(h)-\epsilon_{\sD^{K+1}_{S}}(h)\vert
\end{equation}
where $\sup_{h\in \mathcal{H}}\vert\epsilon_{\sD_{T}}(h)-\epsilon_{\sD^{K+1}_{S}}(h)\vert$ is the upper bound of the task risk gap between the target domain $\sD_{T}$ and the extended domain $\sD^{K+1}_{S}$. Notice $\sD_{S}^{K+1}$ shares the same input space with $\sD_{T}$ since they all use $\{X_{i}^{T}\}_{i=1}^{N_T}$ as inputs. Therefore, we have
\begin{equation}
    d_{\mathcal{H}\Delta\mathcal{H}}(\sD^{K+1}_{S},\sD_{T})=0
\end{equation}
Substituting $(5)$ into $(4)$, we have 
\begin{equation}
    \begin{split}
        \epsilon_{\sD_{T}}(h_{T})\leq  \epsilon_{\sD^{K+1}_{S}}(h_{T}) +&\frac{1}{2} d_{\mathcal{H}\Delta\mathcal{H}}(\sD^{K+1}_{S},\sD_{T})+\\ \sup_{h\in \mathcal{H}}\vert\epsilon_{\sD_{T}}(h)-&\epsilon_{\sD^{K+1}_{S}}(h)\vert
    \end{split}
\end{equation}
Combining $(3)$ and $(6)$, we get the \textbf{Proposition 1}.

\textbf{The learning bound with empirical risk error.} Proposition 1 shows how to relate the extended source domain $\sD^{K+1}_{S}$ and the target domain $\sD_{T}$. Since we use the finite samples to empirically estimate the $\hat{\epsilon}_{\sD^{K+1}_{S}}(h)$ and $\hat{d}_{\mathcal{H}}(\sD^{K+1}_{S},\sD_{T})$ at the training time, We now proceed to give a learning
bound for empirical risk minimization using $N_{T}$ sampled training data.

Following the learning bound \textbf{Lemma 1,5} in \citet{DBLP:journals/ml/Ben-DavidBCKPV10}, for all $0<\delta<1$, with probability at least $1-\delta$, we have:
\begin{equation}
\begin{split}
    \epsilon_{\sD^{K+1}_{S}}(h)\leq \hat{\epsilon}_{\sD^{K+1}_{S}}(h)&+\sqrt{\frac{4}{N_{T}}(d\log\frac{2eN_{T}}{d}+\log\frac{4}{\delta})}\\
    d_{\mathcal{H}\Delta\mathcal{H}}(\sD^{K+1}_{S},\sD_{T})&\leq\hat{d}_{\mathcal{H}\Delta\mathcal{H}}(\sD^{K+1}_{S},\sD_{T})\\&+4\sqrt{\frac{d\log(2N_{T}+\log(\frac{2}{\delta})}{N_{T}}}
\end{split}
\end{equation}
where $d$ is the VC-dimension of model space $\mathcal{H}$. 

Combining $(2)$ and $(7)$, we get the generalization bound for knowledge distillation with the empirical learning error as follows:
\begin{equation}
    \epsilon_{\sD_{T}}(h_{T})\leq \hat{\epsilon}_{\sD^{K+1}_{S}}(h)+\frac{1}{2}\hat{d}_{\mathcal{H}\Delta\mathcal{H}}(\sD^{K+1}_{S},\sD_{T})
    +C_1
\end{equation}
where $C_1$ is a constant as 
\begin{equation}
\begin{split}
     &C_1=\min\{\\
     &\lambda_1+\sqrt{\frac{4}{N_{T}}(d\log\frac{2eN_{T}}{d}+\log\frac{4}{\delta})}+4\sqrt{\frac{d\log(2N_{T}+\log(\frac{2}{\delta})}{N_{T}}},\\
    & \sup_{h\in \mathcal{H}}\vert\epsilon_{\sD_{T}}(h)-\hat{\epsilon}_{\sD^{K+1}_{S}}(h)\vert +\sqrt{\frac{4}{N_{T}}(d\log\frac{2eN_{T}}{d}+\log\frac{4}{\delta})}.\\
    &\}
\end{split}
\end{equation}
\subsection{Appendix C}
\begin{table*}[htbp]
\centering
\begin{tabular}{cc}
\hline
\multicolumn{1}{c|}{Layer} & Configuration                                                       \\ \hline
\multicolumn{1}{c|}{1}     & 2D Convolution with kernel size 5*5 and output feature channels 64  \\ \hline
\multicolumn{1}{c|}{2}     & BatchNorm, ReLU, MaxPool                                            \\ \hline
\multicolumn{1}{c|}{3}     & 2D Convolution with kernel size 5*5 and output feature channels 64  \\ \hline
\multicolumn{1}{c|}{4}     & BatchNorm, ReLU, MaxPool                                            \\ \hline
\multicolumn{1}{c|}{5}     & 2D Convolution with kernel size 5*5 and output feature channels 128 \\ \hline
\multicolumn{1}{c|}{6}     & BatchNorm, ReLU                                                     \\ \hline
\multicolumn{1}{c|}{7}     & Fully connection layer with output channels 10                      \\ \hline
\multicolumn{1}{c|}{8}     & Softmax                                                             \\ \hline
\end{tabular}
\caption{The 3-layers CNN backbone for \textbf{Digit-5}.}
\label{table:backbone}
\end{table*}
\begin{table*}[htbp]
\centering
\begin{tabular}{c|c|c|c|c}
Parameters             & \multicolumn{4}{c}{Benchmark Datasets}                                                            \\ \hline
                       & \textbf{Amazon Review} &\textbf{Digit-5}                              & \textbf{Office-Caltech10}              & \textbf{DomainNet}                   \\ \hline
Data Augmentation    & None  & \multicolumn{3}{c}{Mixup $(\alpha=0.2)$}                                                          \\ \hline
Backbone           & 3-layers MLP      & 3-layers CNN     & \multicolumn{2}{c}{Resnet101 (pretrained = True)}          \\ \hline
Optimizer              & \multicolumn{3}{c}{SGD with momentum = 0.9}                                                       \\ \hline
Learning rate schedule &   \multicolumn{2}{c}{From 0.05 to 0.001 with cosine decay} & \multicolumn{2}{c}{From 0.005 to 0.0001 with cosine decay} \\ \hline
Batchsize           & 50   & 100                                  & 32                            & 50                          \\ \hline
Total epochs           & \multicolumn{4}{c}{40}                                                                            \\ \hline
Communication rounds   & \multicolumn{4}{c}{r=1}                                                                           \\ \hline
Confidence gate        & \multicolumn{3}{c|}{From 0.9 to 0.95}                                & From 0.8 to 0.95            \\ \hline
\end{tabular}
\caption{Implementation details of our KD3A on four benchmark datasets: Amazon Revoew, Digit-5, Office-Caltech10 and DomainNet.}
\label{table:implement}
\end{table*}

\textbf{Proposition 2} \textit{The KD3A bound is a tighter bound than the original bound, if the task risk gap between the knowledge distillation domain $\sD^{K+1}_{S}$ and the target domain $\sD_{T}$ is smaller than the following upper-bound for all source domain $k \in \{1,\cdots,K\}$, that is, $\epsilon_{\sD^{K+1}_{S}}(h)$ should satisfy:}
\begin{equation}
\begin{split}
         \sup_{h\in \mathcal{H}}\vert\epsilon_{\sD^{K+1}_{S}}(h)-\epsilon_{\sD_{T}}(h)\vert &\leq 
        \inf_{h\in \mathcal{H}}\vert\epsilon_{\sD^{K+1}_{S}}(h)-\epsilon_{\sD_{S}^{k}}(h)\vert\\+\frac{1}{2}d_{\mathcal{H}\Delta\mathcal{H}}&(\sD_{S}^{k},\sD_{T})+\lambda_{S}^{k}
\end{split}
\end{equation}

\textbf{Proof:}

Following the Theorem 2 in \citet{DBLP:journals/ml/Ben-DavidBCKPV10}, for each source domain $\sD_{S}^{k}$ and for all $h_{T}\in \mathcal{H}$, we have
\begin{equation}
        \epsilon_{\sD_{T}}(h_{T})\leq \epsilon_{\sD^{k}_{S}}(h_T) +\frac{1}{2}d_{\mathcal{H}\Delta\mathcal{H}}(\sD^{k}_{S},\sD_{T})+\lambda_{S}^{k}
\end{equation}
where $\lambda_{S}^{k}=\min_{h\in \mathcal{H}}\epsilon_{\sD^{k}_{S}}(h)+\epsilon_{\sD_{T}}(h)$ is the optimal task risk of $\sD_{S}^{k}$ and $\sD_{T}$.

The original bound states that for all $h_{T}\in \mathcal{H}$, we have
\begin{equation}
\epsilon_{\sD_{T}}(h)\leq\sum_{k=1}^{K}\bm{\alpha}_{k}\left(\epsilon_{\sD_{S}^{k}}(h)+\frac{1}{2}d_{\mathcal{H}\Delta \mathcal{H}}(\sD_{S}^{k},\sD_{T})\right)+\lambda_0
\end{equation}
where $\lambda_0=\min_{h\in \mathcal{H}}\sum_{k=1}^{K}\alpha_{k}\epsilon_{\sD^{k}_{S}}(h)+\epsilon_{\sD_{T}}(h)$ and we have the following relations between $\lambda_0$ and $\lambda_{S}^{k}$:
\begin{equation}
    \begin{split}
        \lambda_0&=\min_{h\in \mathcal{H}}\sum_{k=1}^{K}\alpha_{k}\epsilon_{\sD^{k}_{S}}(h)+\epsilon_{\sD_{T}}(h)\\
        &\geq \sum_{k=1}^{K}\alpha_{k}( \min_{h\in \mathcal{H}}\epsilon_{\sD^{k}_{S}}(h)+\epsilon_{\sD_{T}}(h))\\
        &=\sum_{k=1}^{K}\alpha_{k}\lambda_{S}^{k}
    \end{split}
\end{equation}

With $(11-13)$, the original bound $(12)$ can be considered as the weighted combination of the source domains. In addition, the KD3A bound is also the combination of the original bound $(12)$ and the knowledge distillation bound $(2)$. Then we get that the KD3A bound is a tighter bound than the original bound if the knowledge distillation bound $(2)$ is tighter than the single source bound $(11)$ for each source domain $\sD_{S}^{k}$, that is, for all source domain $k \in \{1,\cdots,K\}$ and all $h_{T}\in\mathcal{H}$, the knowledge distillation bound should satisfy:
\begin{equation}
\begin{split}
    \epsilon_{\sD^{K+1}_{S}}(h_{T})+\frac{1}{2}d_{\mathcal{H}\Delta \mathcal{H}}(\sD_{S}^{K+1},\sD_{T})
        \\+\min\{\lambda_1,
        \sup_{h\in \mathcal{H}}\vert\epsilon_{\sD^{K+1}_{S}}(h)-\epsilon_{\sD_{T}}(h)\vert\}\\
        \leq \epsilon_{\sD^{k}_{S}}(h_T) +\frac{1}{2}d_{\mathcal{H}\Delta\mathcal{H}}(\sD^{k}_{S},\sD_{T})+\lambda_{S}^{k}
\end{split}
\end{equation}
Since $d_{\mathcal{H}\Delta\mathcal{H}}(\sD^{K+1}_{S},\sD_{T})=0$ and $\lambda_1$ is a constant, the task risk gap $\sup_{h\in \mathcal{H}}\vert\epsilon_{\sD^{K+1}_{S}}(h)-\epsilon_{\sD_{T}}(h)\vert$ should satisfy the following condition for all $h_{T}\in\mathcal{H}$, that is:
\begin{equation}
\begin{split}
    \sup_{h\in \mathcal{H}}\vert\epsilon_{\sD^{K+1}_{S}}(h)-\epsilon_{\sD_{T}}(h)\vert \leq \epsilon_{\sD^{k}_{S}}(h_T)-\epsilon_{\sD^{K+1}_{S}}(h_{T})\\+\frac{1}{2}d_{\mathcal{H}\Delta\mathcal{H}}(\sD^{k}_{S},\sD_{T})+\lambda_{S}^{k}
\end{split}
\end{equation}
Since condition $(15)$ holds for all $h_{T}\in \mathcal{H}$, we have the tighter bound condition as 
\begin{equation}
\begin{split}
         \sup_{h\in \mathcal{H}}\vert\epsilon_{\sD^{K+1}_{S}}(h)-\epsilon_{\sD_{T}}(h)\vert &\leq 
        \inf_{h\in \mathcal{H}}\vert\epsilon_{\sD^{K+1}_{S}}(h)-\epsilon_{\sD_{S}^{k}}(h)\vert\\+\frac{1}{2}d_{\mathcal{H}\Delta\mathcal{H}}&(\sD_{S}^{k},\sD_{T})+\lambda_{S}^{k}
\end{split}
\end{equation}
\begin{table*}[t]
\setlength\extrarowheight{4.5pt}
\centering
\begin{tabular}{c|ccccc|c}
Methods & mt & mm & sv & syn & usps & Avg \\ 
\hline
Oracle &     $99.5_{\pm 0.08}$    &   $95.4_{\pm0.15}$        &     $92.3_{\pm 0.14}$     &   $98.7_{\pm 0.04}$   & $99.2_{\pm 0.09}$ &  $97.0$   \\
Source-only       &   $92.3_{\pm0.91}$      &    $63.7_{\pm0.83}$       &    $71.5_{\pm0.75}$      &   $83.4_{\pm0.79}$    &$90.71_{\pm 0.54}$    &   $80.3$   \\ \hline
 MDAN    & $97.2_{\pm 0.98}$ &  $75.7_{\pm 0.83}$     &     $82.2_{\pm 0.82}$     &   $85.2_{\pm 0.58}$  &  $93.3_{\pm 0.48}$   & $86.7$    \\
 $\text{M}^{3}\text{SDA}$ &   $98.4_{\pm0.68}$      &    $72.8_{\pm 1.13}$       & $81.3_{\pm 0.86}$          &       $89.6_{\pm 0.56}$    &  $96.2_{\pm0.81}$     &   $87.7$  \\ 
\hline
 CMSS  &  $ 99.0_{\pm0.08}  $    &  $  75.3_{\pm 0.57} $      & $88.4_{\pm0.54}  $       &$93.7_{\pm0.21}  $        &$97.7_{\pm0.13}  $    &  $90.8$   \\ 
\hline 
 DSBN$^*$       &  $97.2$       &     $71.6$      &    $77.9$      &    $88.7$       &   $96.1$   &  $86.3$   \\ 
\hline
FADA  &  $91.4_{\pm 0.7}$    &     $62.5_{\pm 0.7}$      &    $50.5_{\pm 0.3}$      &     $71.8_{\pm 0.5}$      &   $91.7_{\pm 1}$    &  $73.6$   \\
\cline{2-7}
FADA$^*$    &  $92.5$    &     $64.5$      &    $72.1$      &     $82.8$      &   $91.7$    &  $80.8$   \\
\cline{2-7}
SHOT & $98.2_{\pm 0.37}$ & $80.2_{\pm 0.41}$ & $84.5_{\pm 0.32}$ & $\mathbf{91.1_{\pm 0.23}}$ & $97.1_{\pm 0.28}$ & $90.2$\\
\cline{2-7}
KD3A$^\text{\dag}$   &   $99.1_{\pm 0.15}$     & $86.9_{\pm 0.11}$      &    $82.2_{\pm 0.26}$     &    $89.2_{\pm 0.19}$        &  $98.4_{\pm 0.11}$   &  $91.2$ \\
\cline{2-7}
KD3A   &   $\mathbf{99.2_{\pm 0.12}}$     & $\mathbf{87.3_{\pm 0.23}}$      &    $\mathbf{85.6_{\pm 0.17}}$     &    $89.4_{\pm 0.28}$        &  $\mathbf{98.5_{\pm 0.25}}$   &  $\mathbf{92.0}$   \\ 
 \hline
\end{tabular}
\caption{UMDA accuracy $(\%)$ on the \textbf{Digit-5}. *: The best results recorded in our re-implementation. \dag: Methods trained without data-augmentation. Our model KD3A achieves $92.0\%$ accuracy and outperforms all other baselines.}
\label{table:digit5}
\end{table*}
\begin{table}[t]
\begin{tabular}{c|ccc|c}
     & Clipart                    & Infograph                  & Painting                   & Avg             \\ \hline
KD3A$^\text{\dag}$ & $69.7_{\pm 0.67}$          & $21.2_{\pm 0.35}$          & $58.8_{\pm 0.66}$          & $48.8$          \\ \hline
KD3A & $\mathbf{72.5_{\pm 0.62}}$ & $\mathbf{23.4_{\pm 0.43}}$ & $\mathbf{60.9_{\pm 0.71}}$ & $\mathbf{51.1}$ \\ \hline
     & Quickdraw                  & Real                       & Sketch                     &                 \\ \hline
KD3A$^\text{\dag}$ & $15.1_{\pm 0.21}$          & $70.4_{\pm 0.54}$          & $57.9_{\pm 0.41}$          & $48.8$          \\ \hline
KD3A & $\mathbf{16.4_{\pm 0.28}}$ & $\mathbf{72.7_{\pm 0.55}}$ & $\mathbf{60.6_{\pm 0.32}}$ & $\mathbf{51.1}$ \\ \hline
\end{tabular}
\caption{The ablation study for data-augmentation strategies on DomainNet.\dag: Methods trained without data-augmentation.}
\label{table:data-aug}
\end{table}
\subsection{Appendix D: Representation Invariant Bounds For KD3A.}
One reviewer argues that the generalization bound in proposition 1 is not rigorous since the optimization process may change the value of $\lambda$. The optimal joint risk $\lambda$ between source and target domain is defined as $\lambda :=\min_{h\in\mathcal{H}}\epsilon_{S}(h)+\epsilon_{T}(h)$. $\lambda$ is based on the hypothesis space $\mathcal{H}$ and is usually intractable to compute. Considering the fixed model backbones are used in in practice (where the hypothesis space $\mathcal{H}$ is implicitly determined), we follow previous works (i.e. Theorem 1 in \citet{DBLP:conf/icml/LongC0J15} and Theorem 2 in \citet{DBLP:conf/nips/ZhaoZWMCG18}) and consider $\lambda$ as a constant. However, we agree with the fact proposed in \citet{DBLP:conf/icml/0002CZG19} (Section 4.1) that optimizing the $\mathcal{H}-$divergence can learn domain invariant representations, but can also change the representation space. This may change the value of $\lambda$. As such, we take the suggestions of the reviewer and replace the original bound with the new bound in \citet{DBLP:conf/icml/0002CZG19}, which utilizes the $\tilde{\mathcal{H}}-$divergence and the constant term $C$. With this upper bound, we propose a new version for our Proposition 1, Theorem 2 and  Proposition 2 as follows:

\textbf{Proposition 1.} Denoting $C_1:=\min\{\E_{\sD_{S}^{K+1}}[\vert f_{S}^{K+1}-f_{T}\vert],$ $\E_{\sD_{T}}[\vert f_{S}^{K+1}-f_{T}\vert]\}$, we have
\begin{equation*}
    \begin{split}
        \epsilon_{\sD_{T}}(h_{T})\leq \epsilon_{\sD^{K+1}_{S}}(h_{T})+d_{\tilde{\mathcal{H}}}(\sD_{S}^{K+1},\sD_{T})\\
        +\min\{C_1,
        \sup_{h\in \mathcal{H}}\vert\epsilon_{\sD^{K+1}_{S}}(h)-\epsilon_{\sD_{T}}(h)\vert\}
        \label{eq:p1}    
    \end{split}
\end{equation*}

\textbf{Theorem 2.} Denoting $C_2:=\sum_{k=1}^{K+1}\alpha_{k}^{CF}\min\{\E_{\sD_{S}^{k}}[\vert f_{S}^{k}-f_{T}\vert],$ $\E_{\sD_{T}}[\vert f_{S}^{k}-f_{T}\vert]\}$, we have 
\begin{equation*}
 \epsilon_{\sD_{T}}(h_{T})\leq \sum_{k=1}^{K+1}\bm{\alpha}_{k}^{\text{CF}}\left(\epsilon_{\sD^{k}_{S}}(h_{T})+d_{\tilde{\mathcal{H}}}(\sD^{k}_{S},\sD_{T})\right)+C_2
\label{eq:t2}
\end{equation*}
\textbf{Proposition 2.}  Denoting $C_{S}^{k}:=\min\{\E_{\sD_{S}^{k}}[\vert f_{S}^{k}-f_{T}\vert],$ $\E_{\sD_{T}}[\vert f_{S}^{k}-f_{T}\vert]\}$, $\forall k$, the tighter condition should satisfy
\begin{equation*}
\centering
\begin{split}
         \sup_{h\in \mathcal{H}}\vert\epsilon_{\sD^{K+1}_{S}}(h)-\epsilon_{\sD_{T}}(h)\vert &\leq 
        \inf_{h\in \mathcal{H}}\vert\epsilon_{\sD^{K+1}_{S}}(h)-\epsilon_{\sD_{S}^{k}}(h)\vert\\+d_{\tilde{\mathcal{H}}}(\sD_{S}^{k},\sD_{T})&  +C_{S}^{k}
\end{split}
\end{equation*}

The proof in Appendix A-C can directly apply to the new bounds. Moreover, KD3A also works on the above new bounds since the  $\tilde{\mathcal{H}}-$divergence can be optimized by minimizing the Batchnorm-MMD distance.
\begin{table}[t]
\begin{tabular}{c|cccc|c}
 Methods    & \textit{Books}                  & \textit{DVDs}                &  \textit{Elec.}            &     \textit{Kitchen}   &Avg.    \\ \hline
 Source-only& $74.4$          & $79.2$          & $73.5$          & $71.4$     & $74.6$ \\
MDAN & $78.6$          & $\mathbf{80.7}$          & $85.4$          & $86.3$     & $82.8$     \\ 
FADA & $78.1$          & $82.7$          & $77.4$          & $77.5$     & $78.9$ \\
 \hline
KD3A & $\mathbf{79.0}$ & $80.6$ & $\mathbf{85.6}$ & $\mathbf{86.9}$ &$\mathbf{83.1}$\\ \hline
\end{tabular}
\centering
\caption{The UMDA performance on Amazon Review dataset.}
\label{table:amazon}
\end{table}
\begin{table*}[!htbp]
\setlength\extrarowheight{4.5pt}
\centering
\begin{tabular}{c|cccc|c}
Methods & A & C & D & W  & Avg \\ 
\hline
Oracle &     $99.7$    &   $98.4$        &     $99.8$     &   $99.7$   &  $99.4$   \\
Source-only       &   $86.1$      &    $87.8$       &    $98.3$      &   $99.0$    &   $92.8$   \\ \hline
 MDAN    & $98.9$ &  $98.6$     &     $91.8$     &   $95.4$   & $96.1$    \\
 $\text{M}^{3}\text{SDA}$ &   $94.5$      &    $92.2$       & $\mathbf{99.2}$          &       $99.5$   &   $96.4$  \\ 
\hline
CMSS  &    $96.0$     &    $93.7$       & $99.3$  &  $99.6$  &  $97.2$  \\
\hline
 DSBN$^*$        &  $93.2$       &     $91.6$      &    $98.9$      &    $99.3$   &  $95.8$   \\ 
\hline
FADA   &  $84.2_{\pm 0.5}$    &     $88.7_{\pm 0.5}$      &    $87.1_{\pm 0.6}$      &     $88.1_{\pm 0.4}$   &  $87.1$   \\
\cline{2-6}
SHOT & $96.4$ & $96.2$ & $98.5$ & $\mathbf{99.7}$ & $97.7$\\
\cline{2-6}
KD3A$^\text{\dag}$ &   $96.0_{\pm 0.07}$     & $95.2_{\pm 0.08}$      &    $97.9_{\pm 0.11}$        &  $99.6_{\pm 0.03}$   &  $97.2$   \\ 
\cline{2-6}
KD3A   &   $\mathbf{97.4_{\pm 0.08}}$     & $\mathbf{96.4_{\pm 0.11}}$      &    $98.4_{\pm 0.08}$        &  $\mathbf{99.7_{\pm 0.02}}$   &  $\mathbf{97.9}$   \\ 
 \hline
\end{tabular}
\caption{UMDA accuracy $(\%)$ on the Office-Caltech10. *: The best results recorded in our re-implementation. \dag: Methods trained without data-augmentation.}
\label{table:office}
\end{table*}
\subsection{Appendix E: The Implementation of BatchNorm MMD}
We have introduced the \textbf{BatchNorm MMD} with the following loss:
\begin{equation}
\begin{split}
    \sum_{l=1}^{L}\sum_{k=1}^{K+1}\bm{\alpha}_{k}
    \big (\Vert\mu(\bm{\pi}_{l}^{T})-\E(\bm{\pi}_{l}^{k})\Vert_2^2
    +\Vert\mu[\bm{\pi}_{l}^{T}]^2-\E[\bm{\pi}_{l}^{k}]^2\Vert_2^2 \big )
\end{split}
\label{eq: batchnormmmd}
\end{equation}
However, directly optimizing the loss (\ref{eq: batchnormmmd}) requires to traverse all Batchnorm layers, which is time-consuming. Inspired by the suggestions of reviewers, we propose a computation-efficient method containing two steps. First, we directly derive the global optimal solution of $\mu(\bm{\pi}_{l}^{T})$ for loss (\ref{eq: batchnormmmd}), that is, $\forall l, 1\leq l\leq L$, the optimal model $h_{\text{op}}^{T}$ on target domain $\sD_{T}$ should satisfy
\begin{equation}
\begin{split}
        \mu_{\text{op}}(\bm{\pi}_{l}^{T}) &= \sum_{k=1}^{K+1}\bm{\alpha}_{k}\E(\bm{\pi}_{l}^{k})\\
        \mu_{\text{op}}[\bm{\pi}_{l}^{T}]^2 &= \sum_{k=1}^{K+1}\bm{\alpha}_{k}\E[\bm{\pi}_{l}^{k}]^2\
\end{split}
\label{eq:batchnorm2}
\end{equation}
Then we calculate the optimal solution from (\ref{eq:batchnorm2}) as  $\{(\mu_{\text{op}}(\bm{\pi}_{l}^{T}),\mu_{\text{op}}[\bm{\pi}_{l}^{T}]^2)\}_{l=1}^L$, directly substitute this solution into every Batchnorm layer of $h^{T}$ and use it as global model. Although this computation-efficient implementation may seem heuristic, we find it practically work and can achieve the same performance as the original maximization step.
\subsection{Appendix F}
\subsubsection{Implementation Details.} 
We perform UMDA on those datasets with multiple domains. During experiments, we choose one domain as the target domain, and use the remained domains as source domains. Finally, we report the average UMDA results among all domains. The code, with which the most important results can be reproduced, is available at Github\footnote{\underline{github.com/FengHZ/KD3A}}. In this section, we discuss the implementation details. Following previous settings \citep{DBLP:conf/iccv/PengBXHSW19}, we use a 3-layer MLP as backbone for Amazon Review, a 3-layer CNN for Digit-5 and the ResNet101 pre-trained on ImageNet for Office-Caltech10 and DomainNet. The details of hyper-parameters are provided in Table \ref{table:implement} and the backbones and training epochs are set to same in all method comparison experiments. In training process, We use the SGD as optimizer and take the cosine schedule to decay learning rate from high (0.05 for Amazon Review and Digit5, and 0.005 for Office-Caltech10 and DomainNet) to zero. 

\textbf{Data augmentations.} Data augmentations are important in deep network training process. Since different datasets require different augmentation strategies (e.g. rotate, scale, and crop), which introduces extra hyper-parameters, we use mixup \citep{DBLP:journals/corr/abs-1710-09412} as a unified augmentation strategy and simply set the mix-parameter $\alpha=0.2$ in all experiments. For fair comparison, we report the results on both conditions, i.e. with/without data-augmentations. The results are shown in Table \ref{table:data-aug},\ref{table:digit5} and \ref{table:office}. The ablation study in data augmentations indicates that mixup strategy can unify different augmentation strategies on different doman adaptation datasets with only one hyper-parameter. Moreover, KD3A can achieve good results even without data-augmentation. 
\subsubsection{Results on Amazon Review, Digit-5 And Office-caltech10.}

In this part, we report the experiment results on \textbf{Amazon Review}, \textbf{Digit-5} and \textbf{Office-Caltech10}. Amazon Review is a sentimental analysis dataset including four domains: Books, DVDs, Electronics and Kitchen Appliances. Digit-5 is a digit classification dataset including MNIST (mt), MNISTM(mm), SVHN (sv), Synthetic (syn), and USPS (up). Office-Caltech10 contains 10 object categories from four domains, i.e. Amazon (A), Caltech (C), DSLR (D). and Webcam (W). \textbf{Note that results are directly cited from published papers if we follow the same setting.} The results on Table \ref{table:amazon}, \ref{table:digit5} and \ref{table:office} show that our KD3A outperforms other UMDA methods and advanced decentralized UMDA methods. Moreover, our KD3A provides better consensus knowledge on the hard domains such as the \textit{MNISTM} domain on the \textbf{Digit-5}, which outperforms other methods by a large margin.
\newpage
\bibliography{main}
\bibliographystyle{icml2021}

\end{document}